\documentclass{article}

\usepackage{microtype}
\usepackage{graphicx}
\usepackage{subfigure}
\usepackage{booktabs} 

\usepackage{hhline}
\usepackage[utf8]{inputenc}
\usepackage{algorithm}
\usepackage{algorithmic}

\usepackage{amsmath}
\usepackage{amsfonts}
\usepackage{mathtools}

\usepackage{flushend}
\usepackage{multirow}
\usepackage{xcolor}
\usepackage{subfigure}
\usepackage{multicol}
\usepackage{enumitem}

\usepackage{hyperref}

\usepackage[accepted]{icml2021}

\icmltitlerunning{MetaTune: Meta-Learning Based Cost Model \\ for Fast and Efficient Auto-tuning Frameworks}

\begin{document}

\twocolumn[
\icmltitle{MetaTune: Meta-Learning Based Cost Model \\ for Fast and Efficient Auto-tuning Frameworks}



\icmlsetsymbol{equal}{*}

\begin{icmlauthorlist}
\icmlauthor{Jaehun Ryu}{postech}
\icmlauthor{Hyojin Sung}{postech}
\end{icmlauthorlist}
\icmlaffiliation{postech}{Pohang University of Science and Technology, Pohang}
\icmlcorrespondingauthor{Hyojin Sung}{hsung@postech.ac.kr}

\icmlkeywords{Machine Learning, ICML}

\vskip 0.3in
]



\printAffiliationsAndNotice{}  

\begin{abstract}
Deep learning compiler frameworks are gaining ground as a more portable back-end for deep learning applications on increasingly diverse hardware. However, they face the daunting challenge of matching performance offered by hand-tuned target-specific libraries. While auto-tuning frameworks with statistical cost models can provide dynamic and efficient code optimization, they suffer from large space exploration and cost model training overheads.
This paper proposes MetaTune, a meta-learning based cost model that more quickly and accurately predicts the performance of optimized codes with pre-trained model parameters. MetaTune encodes convolution kernel codes as structurally similar graphs to facilitate meta-learning, meta-trains a GNN model with a very small input data set, and then predicts optimization parameters for unseen convolution operations with varying sizes and structures during compilation. 
The resulting framework with MetaTune provides 8 to 13\% better inference time on average for four CNN models with comparable or lower optimization time while outperforming transfer learning by 10\% in cross-platform cases.

\end{abstract}


\section{Introduction}

Deep learning models have recently emerged as an application domain that drives innovations in domain-specific compiler technologies. While many deep learning programming frameworks~\cite{tensorflow,pytorch,mxnet} rely on hand-optimized libraries such as NVIDIA cuDNN/cuBLAS or Intel MKL for their back-ends, code-generating compiler frameworks~\cite{tvm,glow,halide} are exploring its potential as a more flexible and portable solution on increasingly diverse and heterogeneous platforms.  

The key challenge for deep-learning compilers lies in generating highly optimized codes with comparable or better performance than hand-tuned libraries. 

To address the limitations of traditional rule-based heuristics that cannot dynamically adapt to input and hardware characteristics,
more data-driven approaches have been proposed to determine optimization strategies~\cite{autotvm,halide,chameleon}. 
These ``auto-tuning'' compilers use statistical cost models to learn the correlation between programs and runtime behaviors from profiling runs and guide the search of ideal optimization configurations.  

However, this approach incurs significant compilation overheads from searching the huge optimization space even for simple convolution kernels and repeatedly profiling optimized codes to train a cost model. The effort to address the issue focused mainly on improving the search algorithms or reducing/accelerating profiling runs~\cite{chameleon,dhakal2020spatial}, while cost models are still freshly trained from hardware measurements of each tensor operation which has a unique optimization space. 
~\cite{autotvm} reused pre-trained cost models for different tensor operations and with transfer learning, but it still requires several hours and hundreds of iterations for cost models to converge, and its effectiveness is task-dependent. 

In a search for a more general and systematic way of reusing knowledge obtained from other tensor operations or platforms to fine-tune the cost model, 
we observed that {\it meta-learning} can enable the cost model to ``learn to learn'' code-performance correlations and very quickly adapt to the optimization space of {\it any} tensor operations with structural similarity.
A performance regression task with code embeddings as input is not a typical meta-learning application and has sparse prior work. However, with a more flexible meta-learning method that works with any model with gradient descent~\cite{maml}, a cost model may predict performance more accurately in much fewer iterations with meta-learned hyperparameters. 

Thus, we propose MetaTune, a meta-learning based cost model for more accurate and efficient auto-tuning frameworks. We implemented MetaTune as an alternative cost model for the auto-tuning framework in TVM~\cite{tvm}. MetaTune includes (1) a feature generator that encodes code templates of convolution kernels as graphs and augments the embeddings to improve structural similarity, and (2) a meta-learning based cost model that meta-trains with performance prediction tasks generated from a small data set during pre-training and then fine-tunes with profiling runs of target kernel during auto-tuning. 

Our experimental results show that MetaTune outperforms existing cost models, regardless of candidate search algorithms used together, and enables fast adaptation in not only cross-operation but also cross-platform scenarios. MetaTune with batch Bayesian optimization search we suggested in the paper improves the inference time by 8\% and 13\% for all evaluated convolution kernels, compared to~\cite{autotvm} and~\cite{chameleon} respectively. Compared to transfer-learning based cost models, the MetaTune model provides faster and better fine-tuning results (10\% on average) on NVIDIA GPU's of different generations than the pre-training target hardware. 

The contributions of the paper are as follows:
\begin{itemize}[leftmargin=*,parsep=0.01in,topsep=0.01in,itemsep=0in]
    \item {\bf Super-graph input augmentation:} Extending graph-based representation of codes in prior work~\cite{xla}~\cite{qast}, we augmented graphs to fit in a common super-graph template and produced input data with higher structural similarity. As a result, the same MetaTune model with template augmentation provides 4.6\% higher inference time on average than the original input. 
    \item {\bf Meta-learning model formulation: } We designed meta-training tasks for performance regression so that the model can be pre-trained for few-shot learning. The result shows that the auto-tuning frameworks with MetaTune jump-start to predict high-performing optimization parameters within the first dozens of iterations. 
    \item {\bf Performance portability: } We showed that MetaTune can seamlessly interface with existing search algorithms and consistently improve autotuning framework's overall efficiency on different GPU platforms. 
    \item {\bf Complete solution: } We implemented a batch Bayesian optimization search algorithm to cater to the needs of MetaTune, which would not require as much space exploration due to pre-training but incurs higher fine-tuning overheads. Our proposed framework with MetaTune and batch Bayesian optimization achieves the lowest inference time for all CNN models evaluated. 
\end{itemize}

In the rest of the paper, we first provide an overview of auto-tuning frameworks in deep learning compilers we assume for MetaTune, and present the design and implementation of the MetaTune model in that context. Then we evaluate the efficiency of MetaTune in various scenarios in the evaluation section and wrap up with related work and conclusions.

\begin{figure*}
\centering
\includegraphics[width=.8\linewidth]{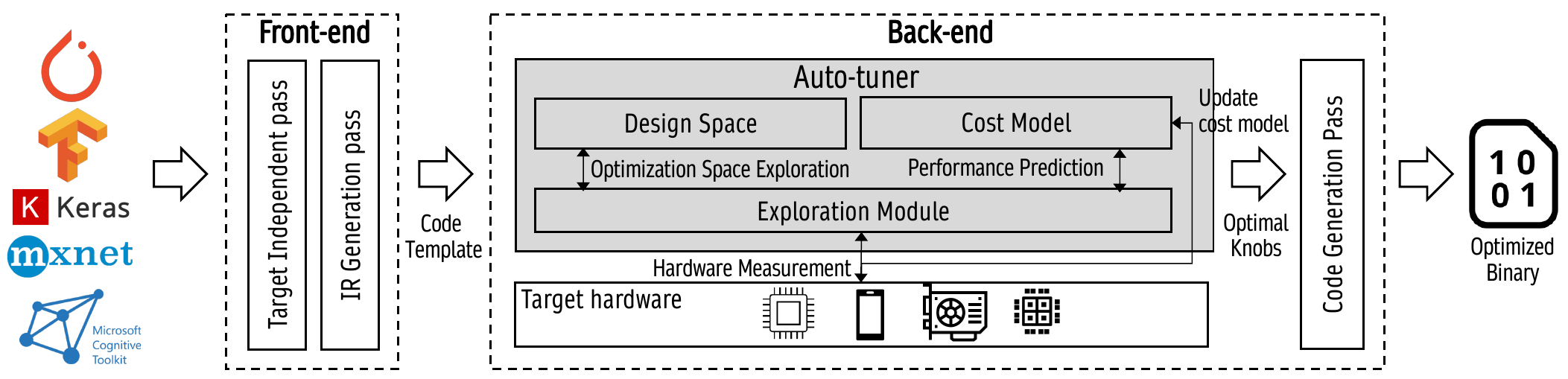}
\vspace{-0.1in}
\caption{The structure of deep learning compiler frameworks}
\label{fig:overview}
\vspace{-0.1in}
\end{figure*}
\section{Background}
\label{sec:background}

\begin{table}
\footnotesize
\centering
\begin{tabular}{p{1.7cm}|l|p{0.6cm}} 
\hline
{\bf Knobs} & {\bf Description}  & {\bf \# of knobs} \\ 
\hline
tile\_x & \multirow{3}{*}{\begin{tabular}[c]{@{}c@{}}Loop tiling parameter on the\\number of filter,height,weight \\ of feature maps \end{tabular}} & 140 \\ 
\cline{1-1}\cline{3-3}
tile\_y &  & 140 \\ 
\cline{1-1}\cline{3-3}
tile\_f &  & 120 \\ 
\hline
tile\_rc & \multirow{3}{*}{\begin{tabular}[c]{@{}c@{}}Loop tile reduction parameter\\ on the number of channels,height\\,weight of feature maps \end{tabular}} & 8 \\ 
\cline{1-1}\cline{3-3}
tile\_rx &  & 2 \\ 
\cline{1-1}\cline{3-3}
tile\_ry &  & 2 \\ 
\hline
\begin{tabular}[c]{@{}c@{}}auto\_unroll\\\_max\_step\end{tabular} & Guide max unroll iteration & 3 \\ 
\hline
unroll\_explicit & Turn on unroll loop & 2 \\
\hline
\end{tabular}
\caption{Description for knobs in TVM}
\vspace{-0.15in}
\label{tab:description_knobs}
\end{table}

\subsection{Deep Learning Compiler Frameworks}

Deep learning compiler frameworks consist of language front-ends and code-generating back-ends, as shown in Figure~\ref{fig:overview}. The front-end takes an input model and translates it into high-level IR (often graph-based) to which target-independent optimizations such as operator fusion and data layout transformation are applied. The back-end takes optimized IR as input and goes through target-dependent optimization passes that further transform IR to better exploit target hardware features. Auto-tuning is implemented as a part of such target-dependent optimization passes.

TVM~\cite{tvm} is an open-source deep learning compilers with the said structure that are widely adopted by industry and academia. Its auto-tuning framework aims to match the performance of hand-tuned libraries and showed promising results in prior work~\cite{autotvm,chameleon}. This paper focuses on proposing an alternative cost model for TVM to further improve auto-tuning efficiency without changing high-level structures, while our approach can be applied to any cost-model based auto-tuning effort.

\subsection{Machine-learning Based Auto-tuning}

The auto-tuning problem, in general, is about automatically generating a search space of optimized codes and finding the best-performing version from hardware measurements~\cite{Autotuning1,Autotuning2}. To efficiently navigate the huge search space of all possible implementations, a cost model function is often used to predict the performance of codes and guide the search. With a code-generating function $\rho$ that uses optimization parameters $\phi$ (e.g., tile size), program operation $\sigma$ (e.g., 2D convolution) and operation options $c$ (e.g., kernel size) to create a search space of optimized codes $D$ and a cost model function $f$ predicts performance of codes in $D$, the problem can be formulated as follows: 

\begin{equation*}
\vspace{-0.2cm}
\phi^*=\operatorname*{argmax}_\phi f(\rho(\phi|\sigma,c))
\end{equation*}
In TVM, $\phi$ consists of a combination of parameters called {\it knobs}, so the above expression turns into finding an optimal combination of knobs as shown in Table ~\ref{tab:description_knobs} to maximize the performance predicted by $f$.

The accuracy of the cost model $f$ is crucial in locating ideal optimization parameters, and many auto-tuning frameworks including TVM adopts machine-learning based cost models that can more dynamically adapt to search results than fixed heuristics or rule-based models. 

As shown in Figure~\ref{fig:overview}, {\it Exploration module} in TVM searches the space of optimized codes for ideal optimization parameters and selects what to profile next based on the predicted performance of {\it the cost model} which is in turn updated based on hardware measurements. The key insight behind MetaTune is that $f$ can be pre-conditioned with randomly sampled points in a small number of similar optimization spaces so that it can quickly adapt to $f^*$, target cost model function.

\begin{figure}[!t]
\begin{center}
\includegraphics[width=0.95\linewidth]{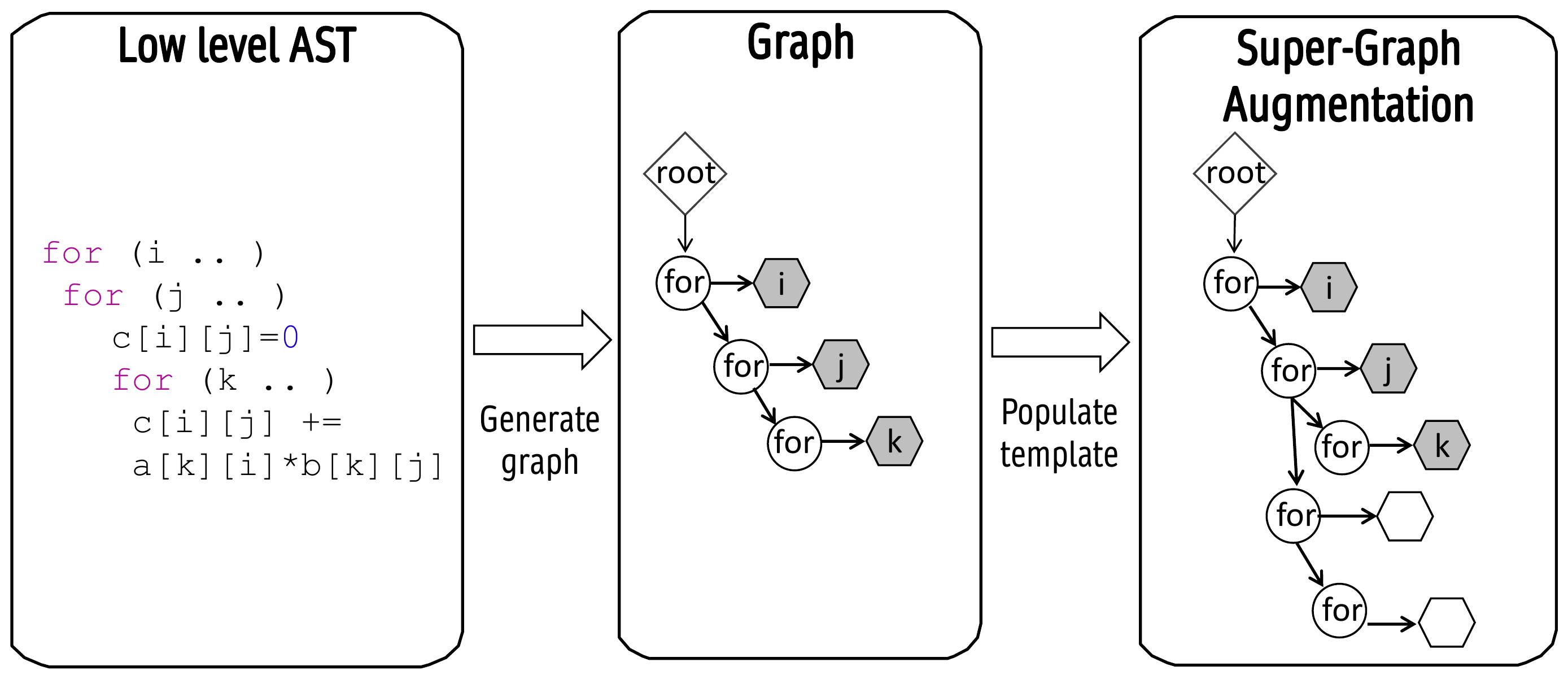}

\caption{Graph-based input generation with super-graph}
\label{fig:template}
\end{center}
\vspace{-0.1in}
\end{figure}

\begin{figure}
\includegraphics[width=\linewidth]{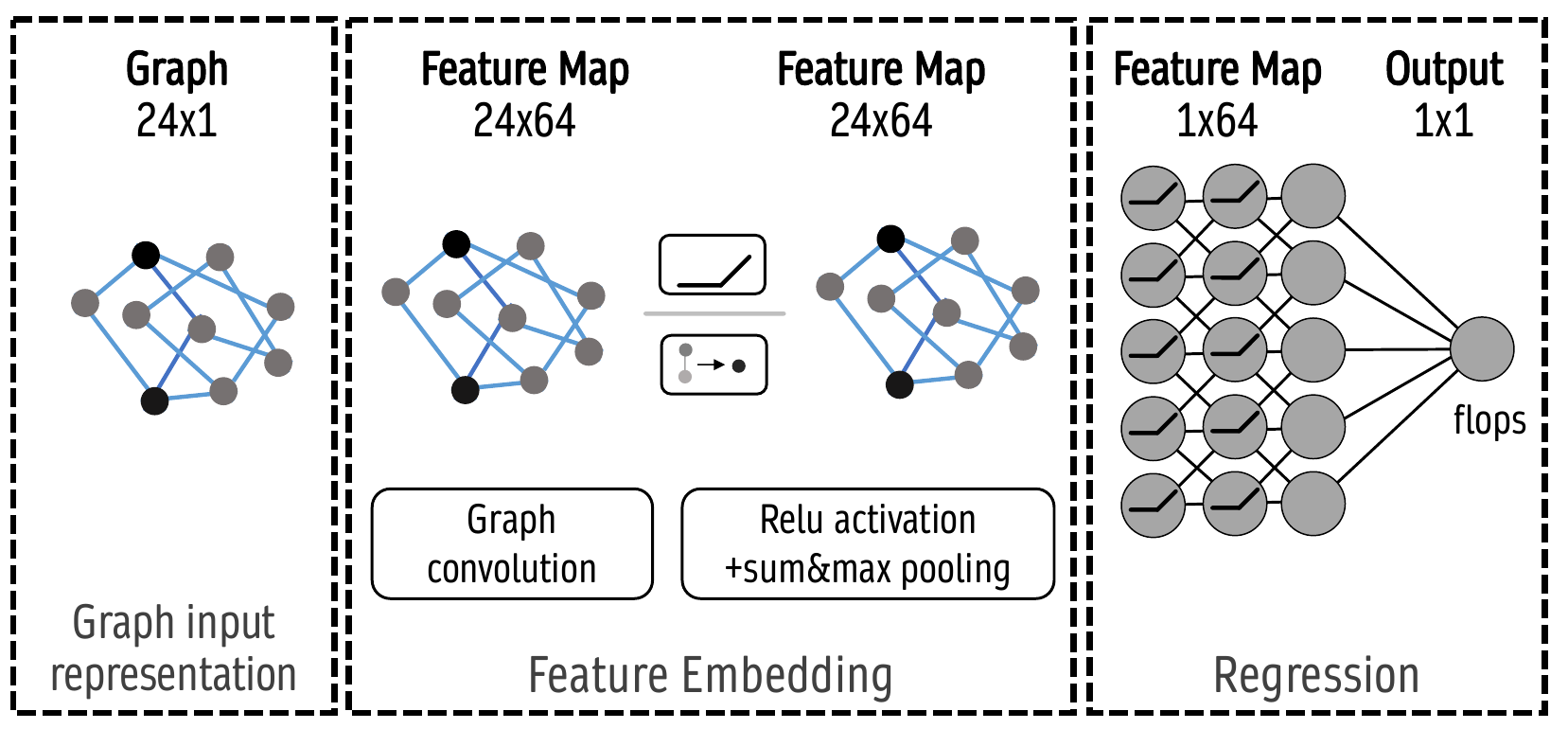}
\vspace{-0.15in}
\caption{MetaTune model architecture}
\label{fig:model}
\vspace{-0.1in}
\end{figure}

\begin{figure*}[hbt!]
\begin{center}
\includegraphics[width=0.85\linewidth]{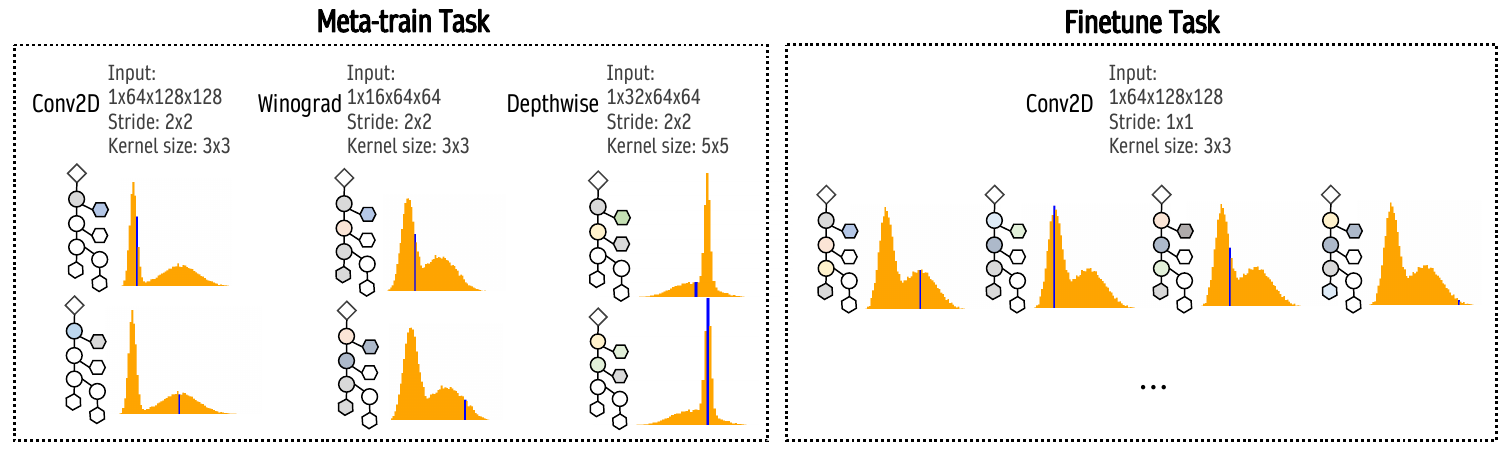}
\vspace{-0.1in}
\caption{Example of few-shot regression tasks for MetaTune. The model learns the universal performance distribution from a pair of graph-structured data which is sampled input and its label (the blue line in the distribution). }
\label{fig:fewshot_learning}
\end{center}
\vspace{-0.1in}
\end{figure*}

\section{Data Representation and Augmentation}
\label{sec:design}

Prior work on automatic optimizations emphasized the importance of finding input features representative of program codes for machine-learning based cost models to accurately understand code-performance correlations~\cite{Cummins2017,autotvm,milepost,TensorComprehensions}. 
When a cost model is meta-learned, it requires input data to be sufficiently similar in terms of its structure as well since the difference in data representation in meta-training and meta-testing dataset can cause meta-overfitting~\cite{maml} leading to a learning failure. For example, meta-learning an image-based model assumes input images of the same size where values for each pixel are in the same value range. 
Thus, we focus on designing input data representation of MetaTune to have to enable meta-learning.

\subsection{Graph Representation} 

TVM offers three ways to generate input data for the cost model: a set of knob parameters, ``loop context vectors'' which include computed statistics about loops such as loop lengths, access patterns, and many different arithmetic instructions, and curve features which summarize relational information about loop context vectors~\cite{autotvm}. 
Our attempt to design a meta-learning based cost model with these representations consistently failed due to a lack of structure in them.

They express structural information as second-hand statistics computed and summarized from codes. 
As weakly structured data, they have variable lengths for tensor operations with different structures. 
Such format divergence interfered with meta-learning adaptation which assumes structurally similar tasks. Inspired by prior work that showed more structured, tree/graph-based code representation improves the modeling efficiency in general~\cite{kaufman2020learned,qast,cogr}, we use graph-based input data for meta-learning and fine-tuning the MetaTune cost model.

Convolution kernels are expressed as code templates in TVM where kernels of the same type (e.g., \texttt{conv1d}, \texttt{conv2d+transponse}) have the same code template. Code templates are in turn represented as low-level abstract syntax trees (AST).

As shown in Figure~\ref{fig:template}, we extract loop structures by recursively traversing the AST and build a summarized graph representation in the middle. Nodes in the graph can be a \texttt{root} node, \texttt{for} nodes, or \texttt{iterval} nodes. Directed edges between the \texttt{root} node and all \texttt{for} nodes represent control flow, while edges between \texttt{for} and \texttt{iterval} nodes match loops with detailed loop metadata. Each \texttt{iterval} node has loop context vectors computed by TVM as its node feature. \texttt{root} and \texttt{for} nodes, without any node features, serve the role of placing the loop context vectors in a precise context.

\subsection{Super-graph Augmentation}

To produce more structurally uniform graphs, we augment the AST-based graph input to fit in the shared super-graph template. 
This augmentation can be viewed as a further format standardization of code templates (which are already per-type format standardization) across {\it all} types of convolution operations.

We perform a union of all possible graph representations and create a super-graph template with placeholder nodes and edges between them. For super-graph augmentation, we locate a corresponding \texttt{iterval} node in the template which is uniquely identified and populate the node with context feature vectors. Figure~\ref{fig:template} shows that only  $i$, $j$, and $k$ \texttt{iterval} nodes corresponding to the nodes in the original graph contain context feature vectors, while other \texttt{iterval} nodes are null in the super-graph version on the right. To save the overhead of augmentation, we pre-define the super-graph template for all convolution kernels supported in TVM, identify mappings between matching nodes in a table, and use the table to quickly generate final output.
We plan to extend the super-graph formulation process in the future to dynamically support a wider range of tensor operations.

\section{Model Architecture}
\label{sec:model}

The MetaTune cost model consists of a graph convolutional network (GCN)~\cite{gcn}, an activation and aggregation layer, and three fully connected (FC) layers with ReLU activation, as shown in Figure~\ref{fig:model}. 

The GCN model serves as an embedding layer that takes augmented graphs as input data and encodes relationships between performance-critical information through convolutions on graph nodes. 
With the following ReLU activation and weighted sum and max aggregation, low-dimensional, fixed-length embedding vectors are generated as input for meta-training and fine-tuning tasks. 
The regression model uses fully connected layers with non-linearity as exemplified in~\cite{maml} for regressions tasks. 
This model is meta-trained with the model-agnostic meta-learning (MAML) method and continues to fine-tune with inference queries during auto-tuning. The output of the model is the predicted performance of input implementation in FLops.

\section{Meta-learning Process}
\label{sec:process}
\begin{algorithm}[t]
\caption{Metatune Meta Train-step}
\begin{algorithmic}[1]
\REQUIRE Input-Performance dataset $\mathcal{D}$ and distribution $p(\mathcal{T})$ over kernel performance
\REQUIRE $\alpha$, $\beta$: step size
\REQUIRE $n$,$\gamma$ : epochs, learning rate
\STATE Randomly initialize parameters $\theta$

\FOR{i \dots $n$}

    \STATE Sample input,output pair($knob,perf$) in $\mathcal{D}$ 
    \STATE Evaluate training loss $\nabla_\theta \mathcal{L}(f_{\theta}(knob),perf)$ 

    \STATE Update $\theta \leftarrow \theta - \gamma\nabla_\theta \mathcal{L}(f_{\theta}(knob),perf))$
\ENDFOR  

\WHILE{not done} 

\STATE Sample batch of tasks $\mathcal{T}_i \sim p(\mathcal{T})$
  \FORALL{$\mathcal{T}_i$}
 \STATE Evaluate training loss $\nabla_\theta \mathcal{L}(f_{\theta}(knob_i),perf_i)$ 
 \STATE Compute adapted parameters with gradient descent:\\ $\theta_i'=\theta-\alpha \nabla_\theta \mathcal{L}(f_{\theta}(knob_i),perf_i)$
 \ENDFOR
 \STATE Update $\theta \leftarrow \theta - \beta \nabla_\theta \sum_{\mathcal{T}_i \sim p(\mathcal{T})}  \mathcal{L}( \theta_i')$
\ENDWHILE

\end{algorithmic}
\label{alg:metatune}
\vspace{-0.01in}
\end{algorithm}

We trained our cost model using MAML~\cite{maml,simaese} as shown in Algorithm~\ref{alg:metatune}.

\begin{enumerate}[leftmargin=*,parsep=0.01in,topsep=0.01in,itemsep=0in]
    \item {\bf Supervised pre-training} [line 1-5]:  As previous work~\cite{mzsr} on regression models learned with meta-learning proposed, we prepare a small dataset with labels and pre-train the GCN embedding layer via supervised learning to obtain stable embedding results from GCN and model weights that are close to a target cost model function $f^*$.
    \item {\bf Input data preparation for meta-training} [line 8]: We prepare another labeled dataset for all types of convolution kernels with varying configurations (more details on training dataset in Section~\ref{sec:method}) and randomly organized them into meta-training tasks. 
    
    We create few-shot meta-training tasks to approximate a universal performance distribution for knobs for all convolutional kernels. For the example shown in Figure~\ref{fig:fewshot_learning}, a 3-way, 2-shot learning task includes two examples with different optimization parameters from each of three classes for \texttt{conv2d}, \texttt{winograd}, and \texttt{depthwise} operations with specific input and kernel size and stride.

    \item {\bf Meta-training} [line 9-13]: The regression model with FC layers learns the meta-training tasks with MAML (in this phase, we only use the pre-trained GCN layer without further training). 
    For each meta-training task, the model computes loss (line 11) and updates $\theta_i$ with gradient descent (line 12), where $\alpha$ is the learning rate for local parameter updates within a task.
    After all tasks are evaluated, $\theta$ of the cost model $f$ is updated with gradient descent with the global learning rate $\beta$ and the sum of $\theta_i$ (line 13).
    
    The example in Figure~\ref{fig:fewshot_learning} will sample a batch of tasks in each iteration and trains the model with the training loss between predicted FLops and actual FLops from each task.  
    The model will learn to predict the performance of an unseen implementation of the same task after being trained with other graphs in the task through many epochs. At the end of training, the model will be in a state (in terms of model weights) to need only a small number of samples to regress a structurally similar but possibly unseen task.
    
\end{enumerate}

During compilation, the pre-trained MetaTune model keeps improving with adaptation to inference queries from the exploration module. Unlike meta-training tasks, fine-tuning tasks include candidate samples for a single task to be optimized. Online hardware measurement will be used for gradient updates.

\section{Evaluation Methodology}
\label{sec:method}

\begin{table}
\centering
\footnotesize
\begin{tabular}{p{1.45cm}|p{1.3cm}|p{1.5cm}|p{1.5cm}} 
\hline
\bf{Operation}                                                                & \bf{Input size} & \bf{Input channel} & \bf{Output channel}  \\ 
\hline
\begin{tabular}[c]{@{}l@{}}conv1d,\\transpose1d\end{tabular}           & 150-600     & 32-128         & 32-512           \\ 
\hline
\begin{tabular}[c]{@{}l@{}}conv2d,\\transpose2d,\\winograd\end{tabular} & 7-224       & 3-128          & 16-128           \\
\hline
\end{tabular}
\vspace{-0.05in}
\caption{Parameters for MetaTune model training dataset}
\label{table:Training}
\vspace{-0.1in}
\end{table}

We implemented MetaTune as a plug-in cost model component for TVM version 0.7.dev. 
We replaced the input feature generator and cost model with MetaTune implementation while experimenting with different search algorithms in the exploration module as follows.

\noindent

\begin{itemize}[leftmargin=*,parsep=0in,topsep=0in]
      
    \item Batch Bayesian optimization:  
      
        The default search algorithm in TVM, simulated annealing (SA)~\cite{sa}, is known to effectively approximate global optimization in a large search space with extensive random exploration. 
        We observed that MetaTune would not require as much space exploration but incur higher inference overheads due to a complicated model structure than prior work, thus implemented a batch Bayesian optimization (BO) algorithm inspired by~\cite{Desautels:JMLR:2014,qucb,pbo}. The BO is more exploitation-centric and performs a search with a much lower number of iterations than SA. 
        Our implementation using \texttt{botorch} library defines each knob to be the input of the BO to understand the knob relationships and executes all BO operations and inference queries in parallel on GPU. 
 
    \item Adaptive exploration and sampling~\cite{chameleon}: Reinforcement-learning based search algorithms. 
  
\end{itemize}

\vspace{0.1in}
\noindent
{\bf Evaluated models and data: }
We evaluated inference times, total optimization times, cost model prediction accuracy (MSE) for four CNN models: \texttt{ResNet-18},  \texttt{Vgg-16}, \texttt{Squeezenet\_v1.0}, and \texttt{AlexNet}. 

For meta-training, we created a meta-training dataset from 47 convolution kernels as shown in Table~\ref{table:Training}. Their input size and input/output channels are randomly chosen within the range in the table, while the stride and padding were fixed at 3 and 1 respectively. 
Then we randomly extracted 200 samples of optimized codes for each class, from which meta-training tasks are created. 

\vspace{0.1in}
\noindent
{\bf Evaluated frameworks: }
We evaluated the following seven auto-tuning frameworks in TVM including reproduction of prior work~\cite{autotvm, chameleon}. 
\begin{itemize}[leftmargin=*,parsep=0.01in,topsep=0.01in,itemsep=0in]
    \item {\bf xgb: }The default TVM implementation with gradient boosting~\cite{xgb} for the cost model and SA for search algorithm as in~\cite{autotvm}. 
    \item {\bf xgb-Xfer: } {\bf xgb} with transfer learning enabled (as in TVM v0.7.dev).     
    \item {\bf RL: } The reinforcement-learning based implementation in~\cite{chameleon}.
    \item {\bf meta-RL} and {\bf meta-RL-T: } MetaTune with RL-based candidate selection. {\bf meta-RL-T} and {\bf meta-RL} work with and without super-graph augmented data respectively.
    \item {\bf meta-BO} and {\bf meta-BO-T: } MetaTune with batch Bayesian optimization based candidate selection. {\bf meta-BO-T} and {\bf meta-BO} work with and without super-graph augmented data respectively. 
\end{itemize}

\vspace{0.1in}
\noindent
{\bf Evaluation environment: }
We conducted our primary performance evaluation for all frameworks on a server with Intel Xeon CPUs and NVIDIA geForce 2080 Ti GPUs. Hardware measurements, batch Bayesian optimization, and RL-based implementation use a single GPU, while other components run on CPU. 
For experiments on additional target hardware, we used an NVIDIA Titan XP GPU server and an AMD Radeon RX VEGA 64 GPU server.

\section{Experimental Results}
\label{sec:result}

Our experimental evaluation aims to answer the following questions: Can MetaTune consistently adapt faster and find better solutions with different search algorithms on different platforms? 
We answered the question by (1) measuring inference time of CNN models optimized by auto-tuning frameworks with MetaTune (evaluating the overall framework efficiency), (2) measuring the model accuracy of MetaTune (isolating the cost model efficiency), and (3) repeating the experiments on other platforms (showing portability). 
We performed hyperparameter tuning for the MetaTune model with batch Bayesian optimization and RL-based candidate search algorithms, and the details can be found in the appendix. 

\begin{figure*}[!htb]

\subfigure[Resnet-18]{
\includegraphics[width=.23\textwidth]{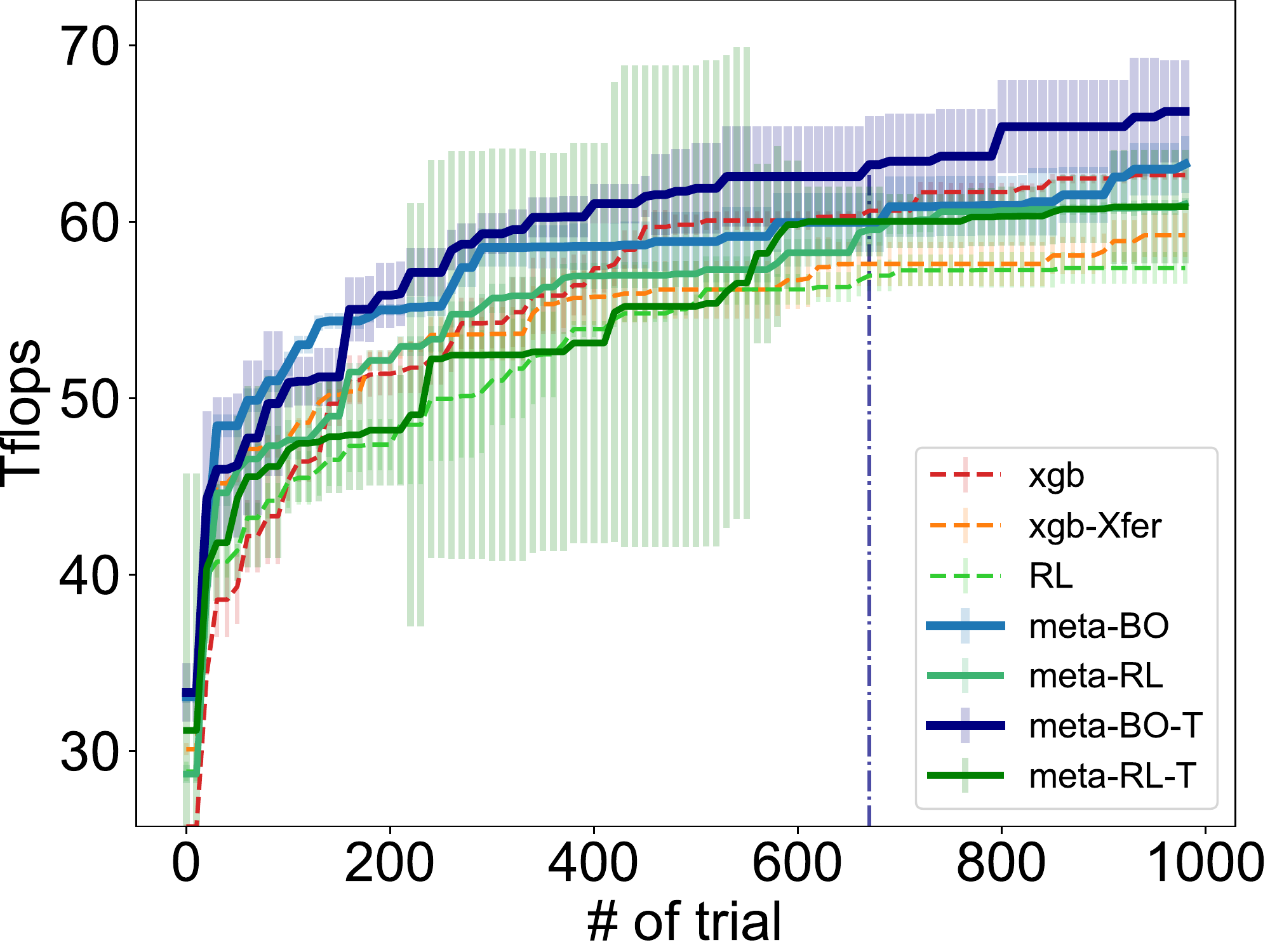}
\label{fig:ti:trial:resnet-18}
}
 \hfill
\subfigure[Vgg-16]{
\includegraphics[width=.23\textwidth]{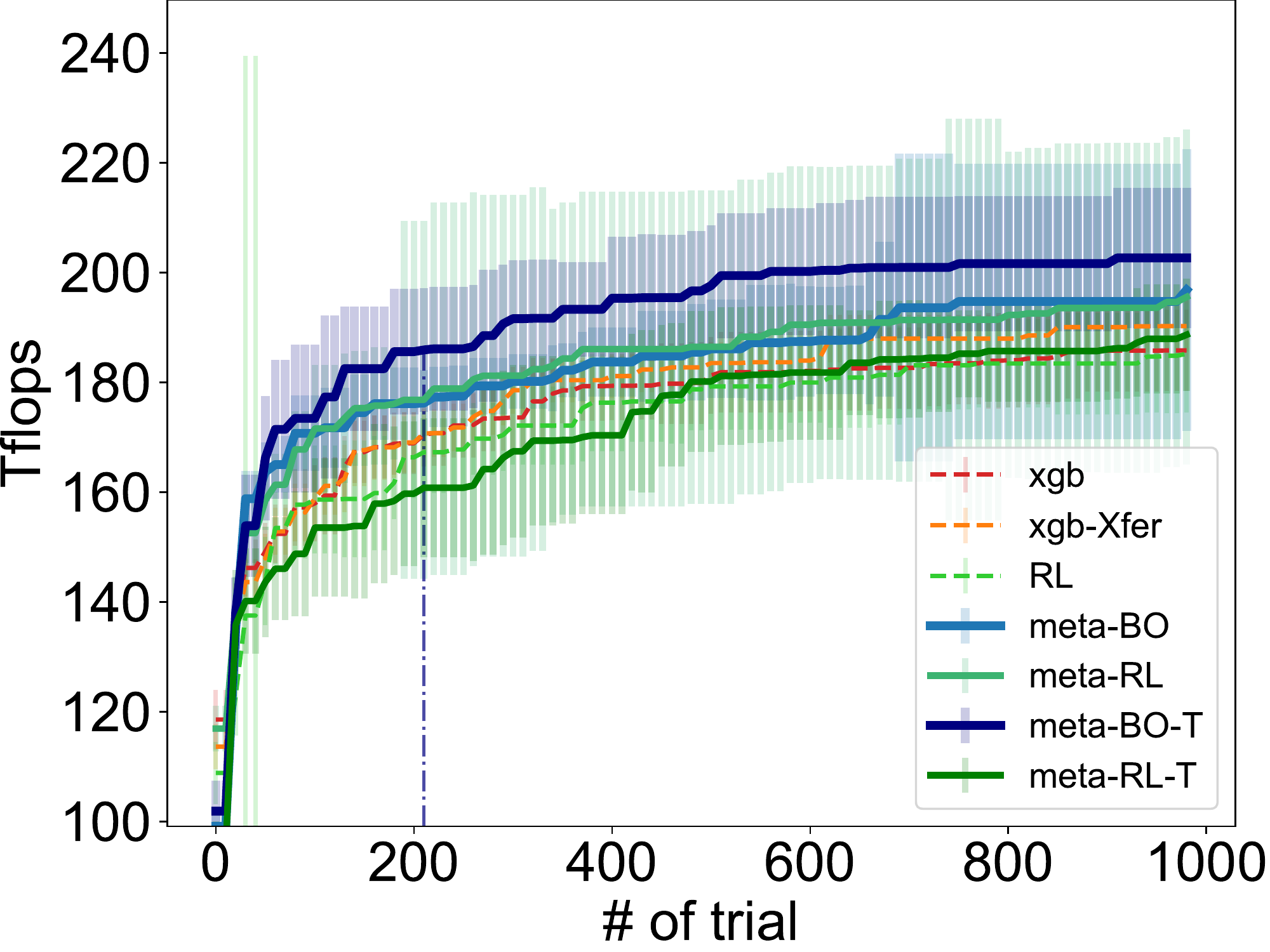}
\label{fig:ti:trial:vgg-16}
}
\subfigure[Squeezenet]{
\includegraphics[width=.23\textwidth]{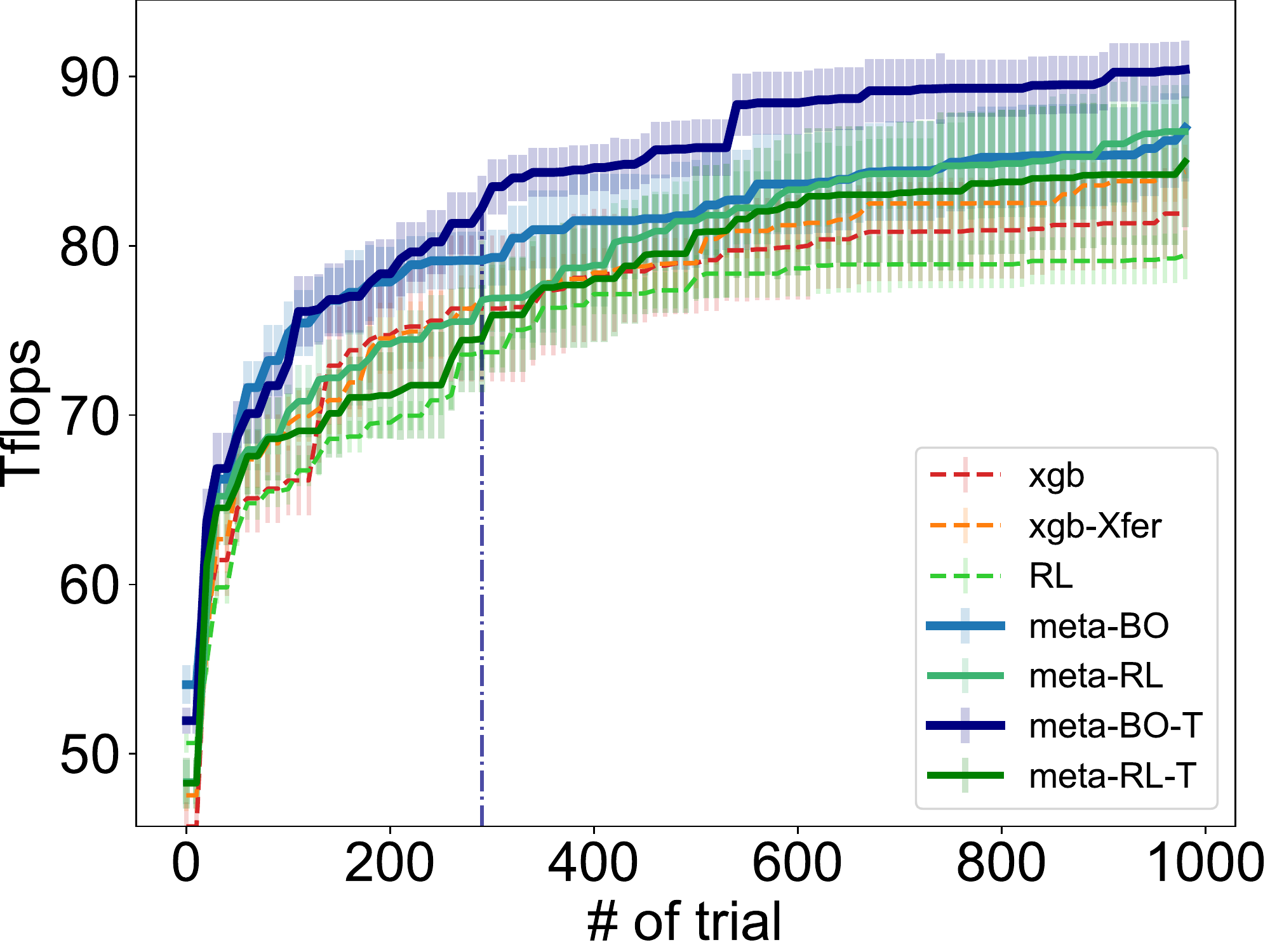}
\label{fig:ti:trial:squeezenet_v1.0}
}
 \hfill
\subfigure[Alexnet]{
\includegraphics[width=.23\textwidth]{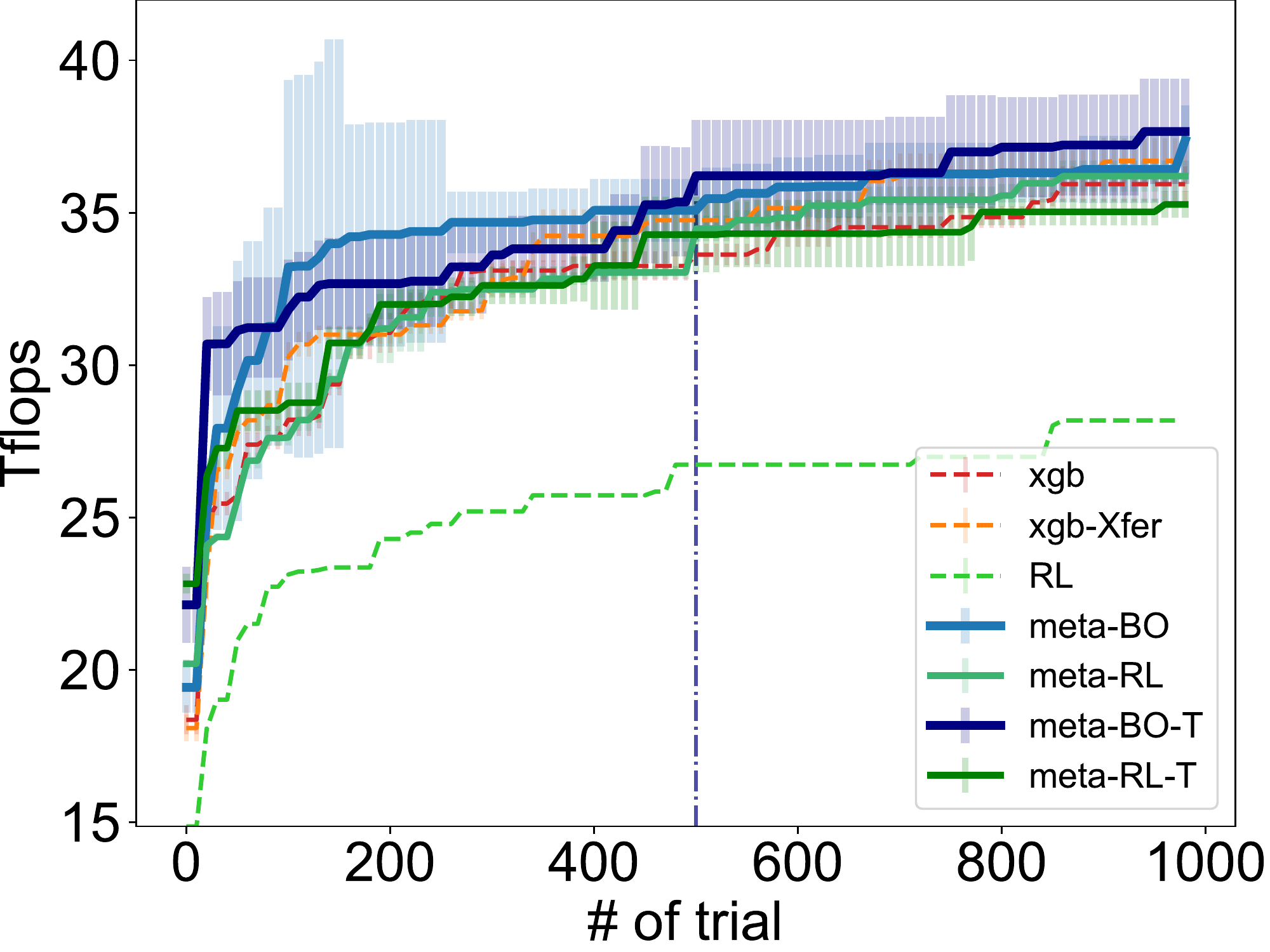}
\label{fig:ti:trial:alexnet}
}
\vspace{-0.1in}
\caption{Output performance}
\label{fig:ti:trial:result}
\vspace{-0.1in}
\end{figure*}

\begin{figure*}[!htb]

\subfigure[Resnet-18]{
\includegraphics[width=.23\textwidth]{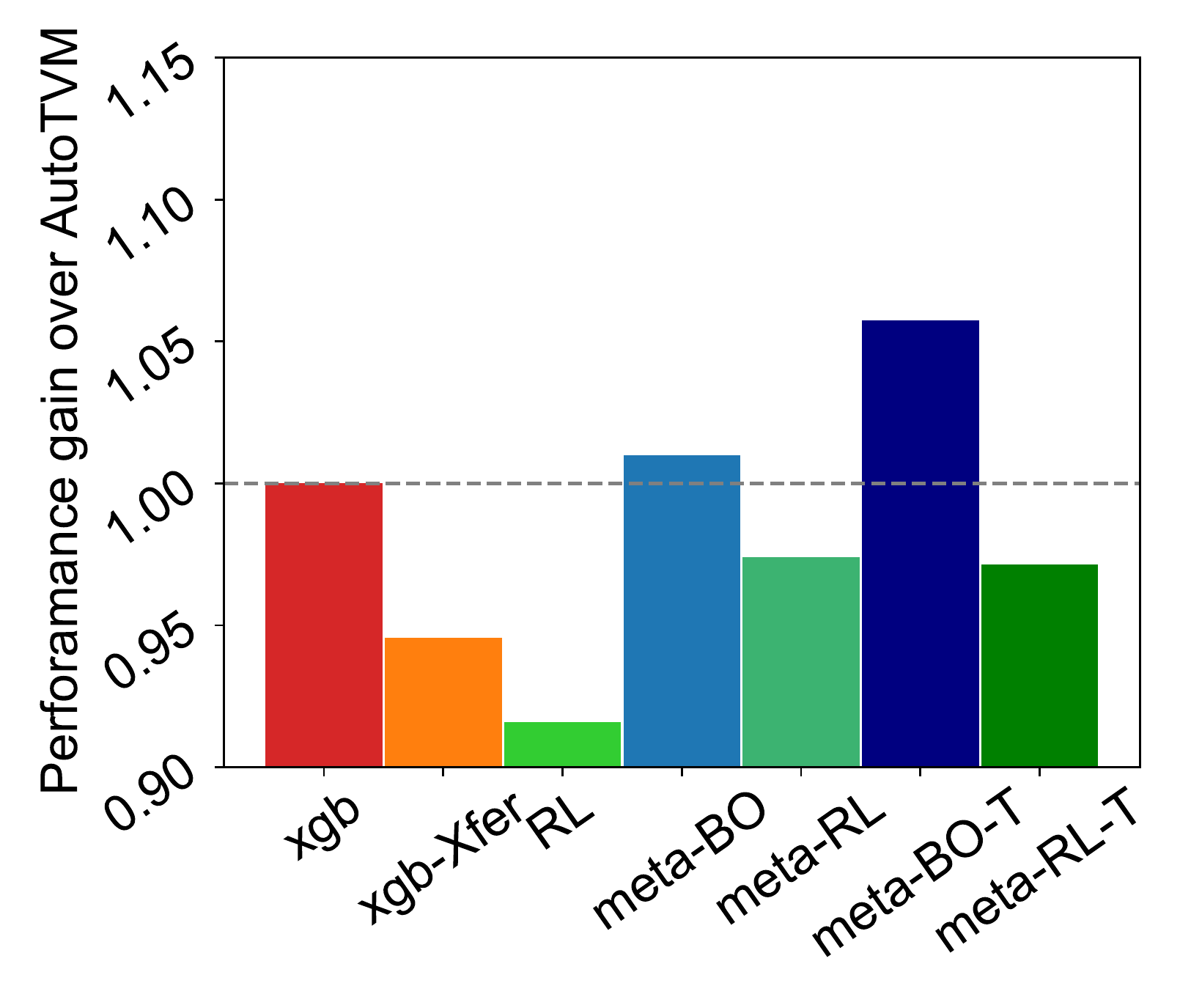}
\label{fig:ti:bar:resnet-18}
}
 \hfill
\subfigure[Vgg-16]{
\includegraphics[width=.23\textwidth]{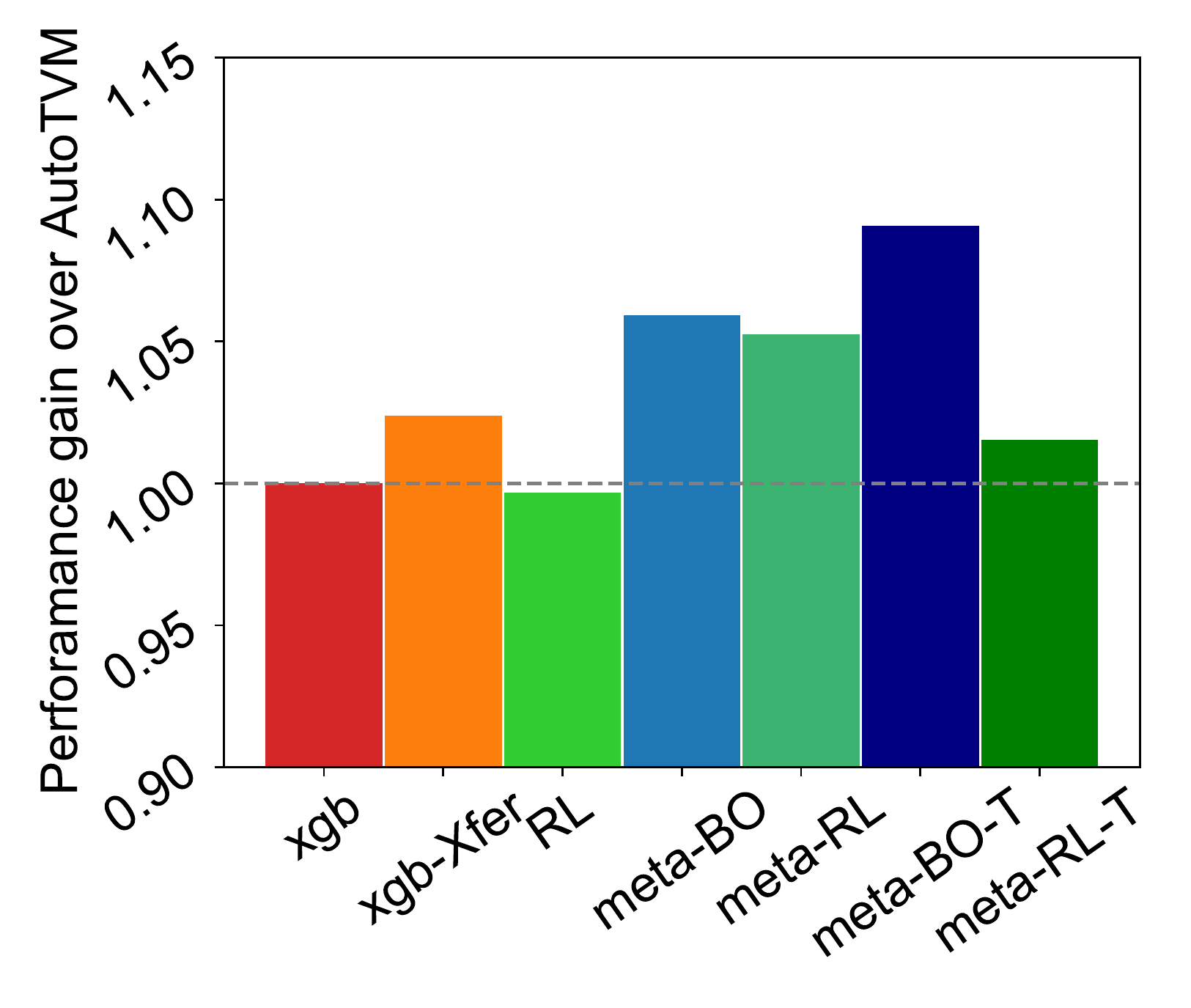}
\label{fig:ti:bar:vgg-16}
}
\subfigure[Squeezenet]{
\includegraphics[width=.23\textwidth]{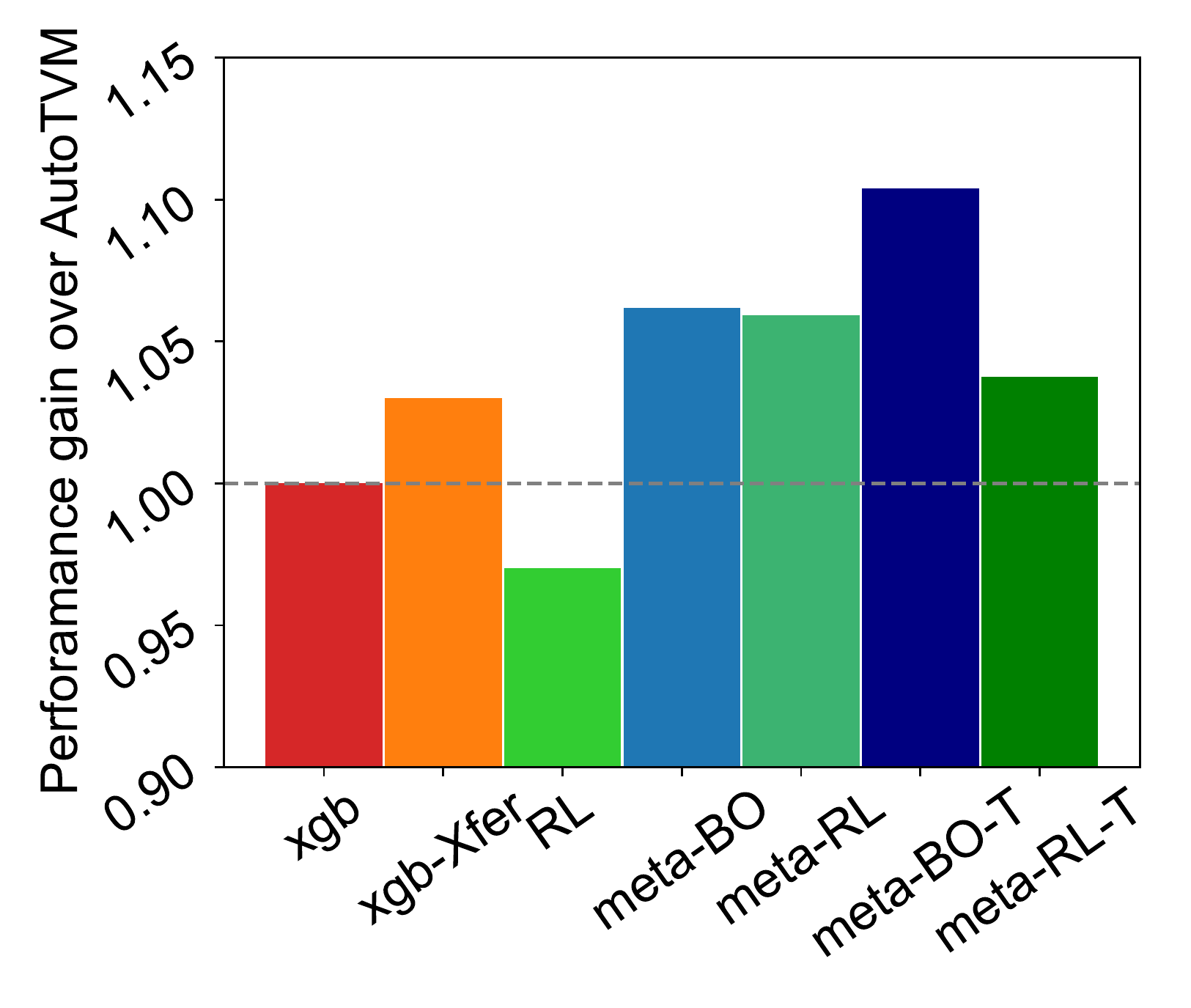}
\label{fig:ti:bar:squeezenet_v1.0}
}
 \hfill
\subfigure[Alexnet]{
\includegraphics[width=.23\textwidth]{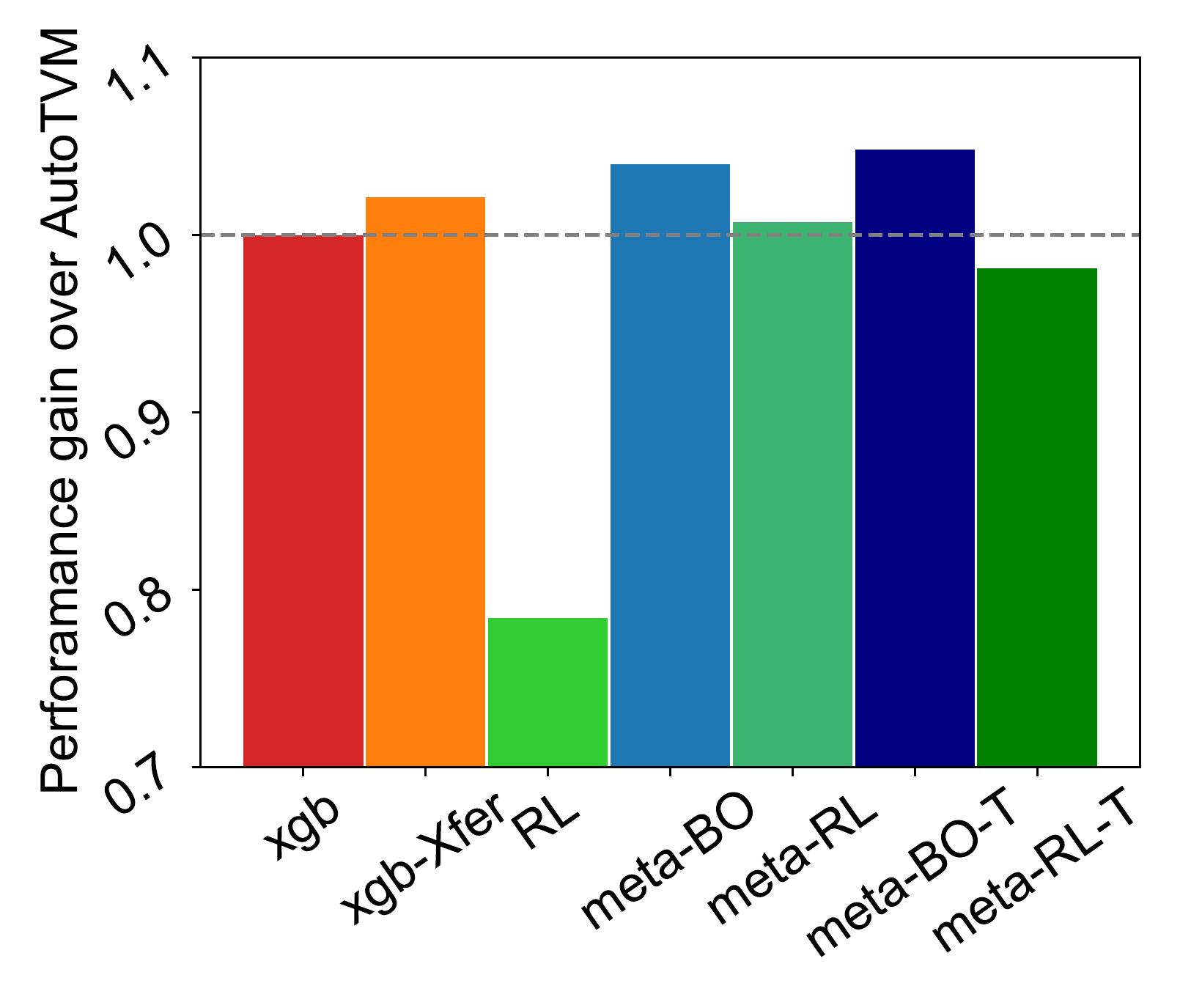}
\label{fig:ti:bar:alexnet}
}
\vspace{-0.1in}
\caption{End-to-end evaluation of normalized inference time}
\label{fig:ti:bar:result}
\vspace{-0.1in}
\end{figure*}

\begin{figure*}[!htb]
\centering
\begin{minipage}[b]{.45\textwidth}
\subfigure[Resnet-18]{
\includegraphics[width=.47\textwidth]{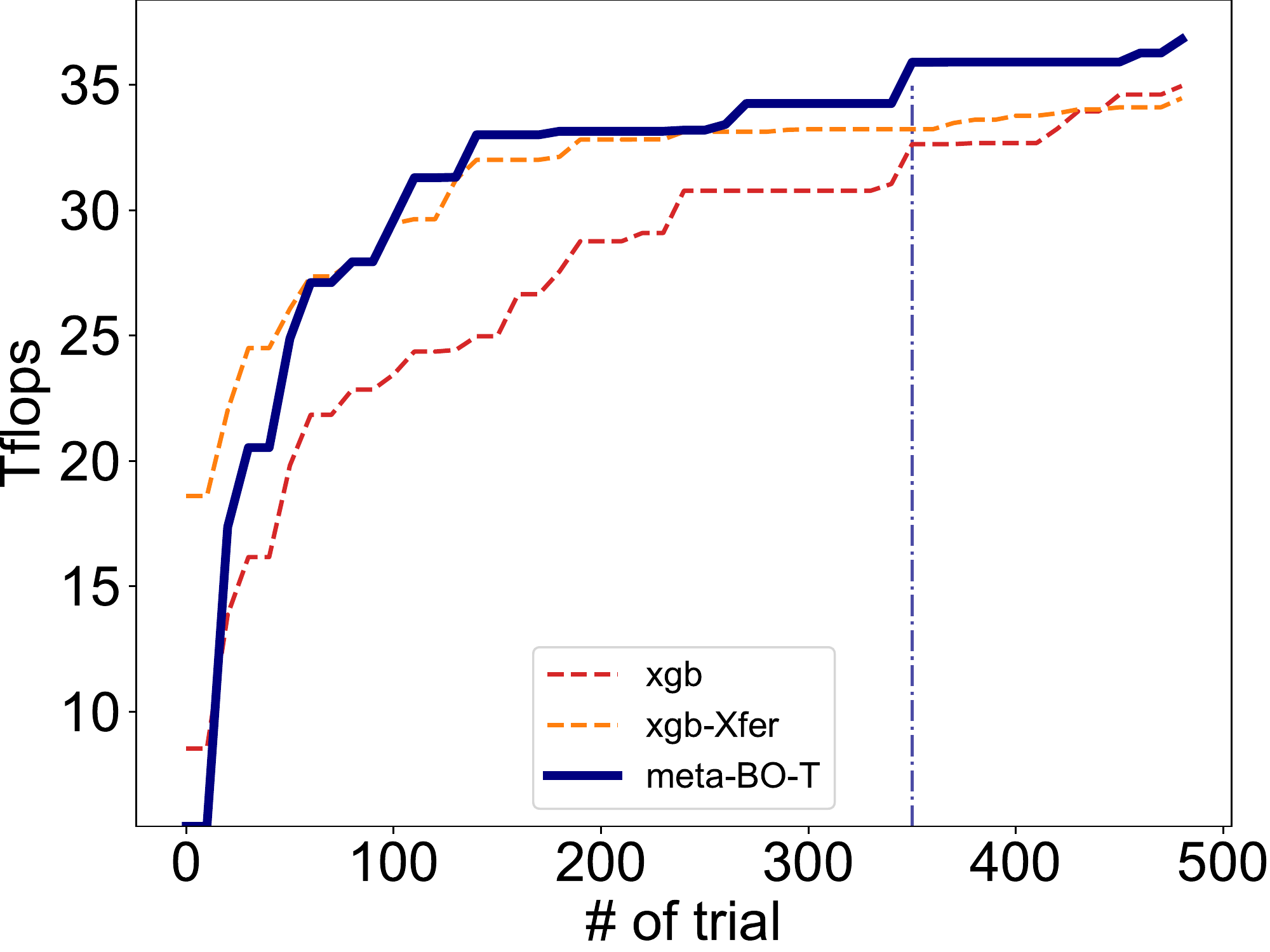}
\label{fig:TITAN:resnet-18}
}
\subfigure[Alexnet]{
\includegraphics[width=.47\textwidth]{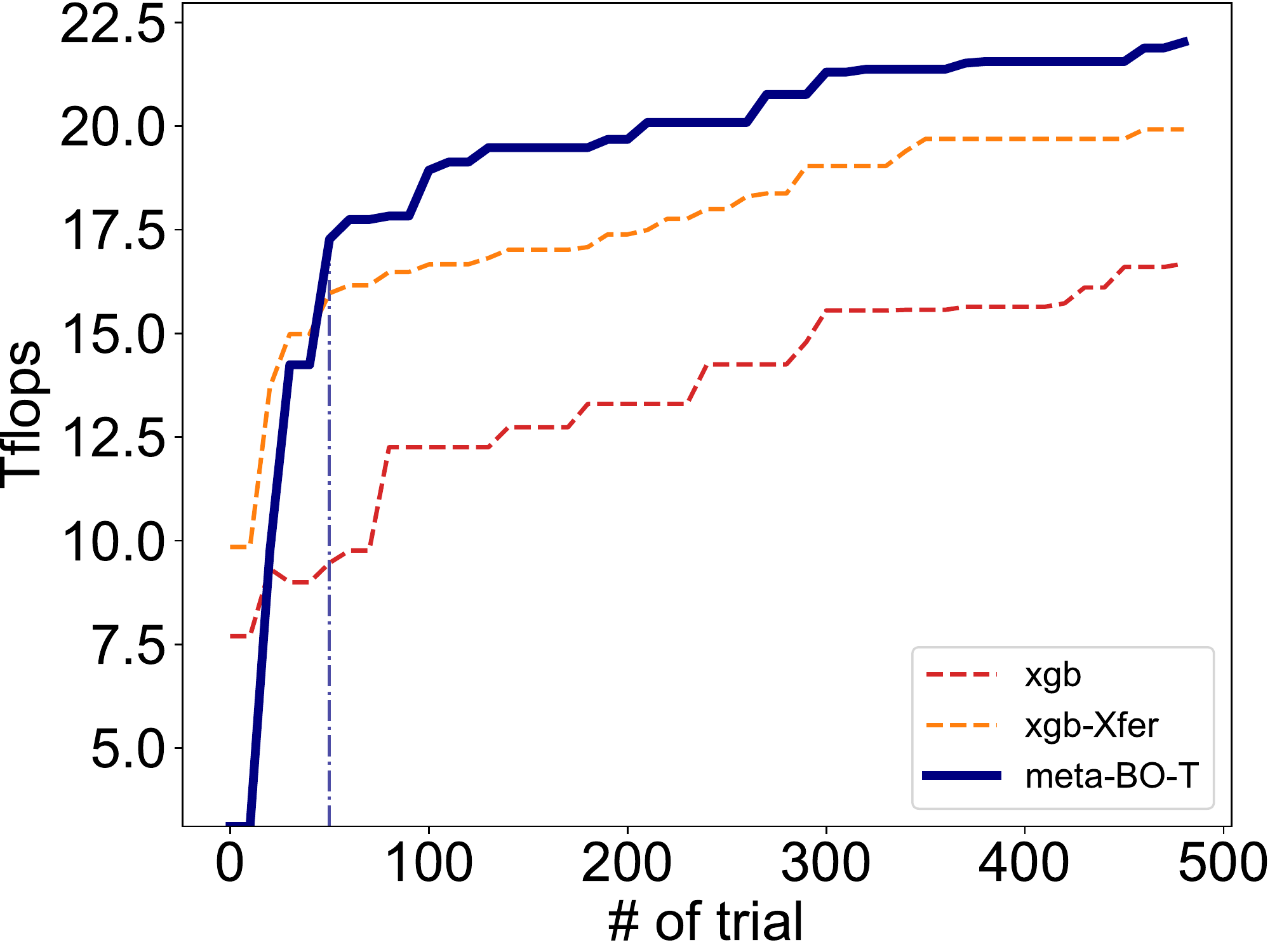}
\label{fig:TITAN:alexnet}
}
\caption{Output performance on NVIDIA Titan XP GPU}
\label{fig:TITAN:result}
\end{minipage}\qquad
\begin{minipage}[b]{.45\textwidth}
\subfigure[Resnet-18]{
\includegraphics[width=.47\textwidth]{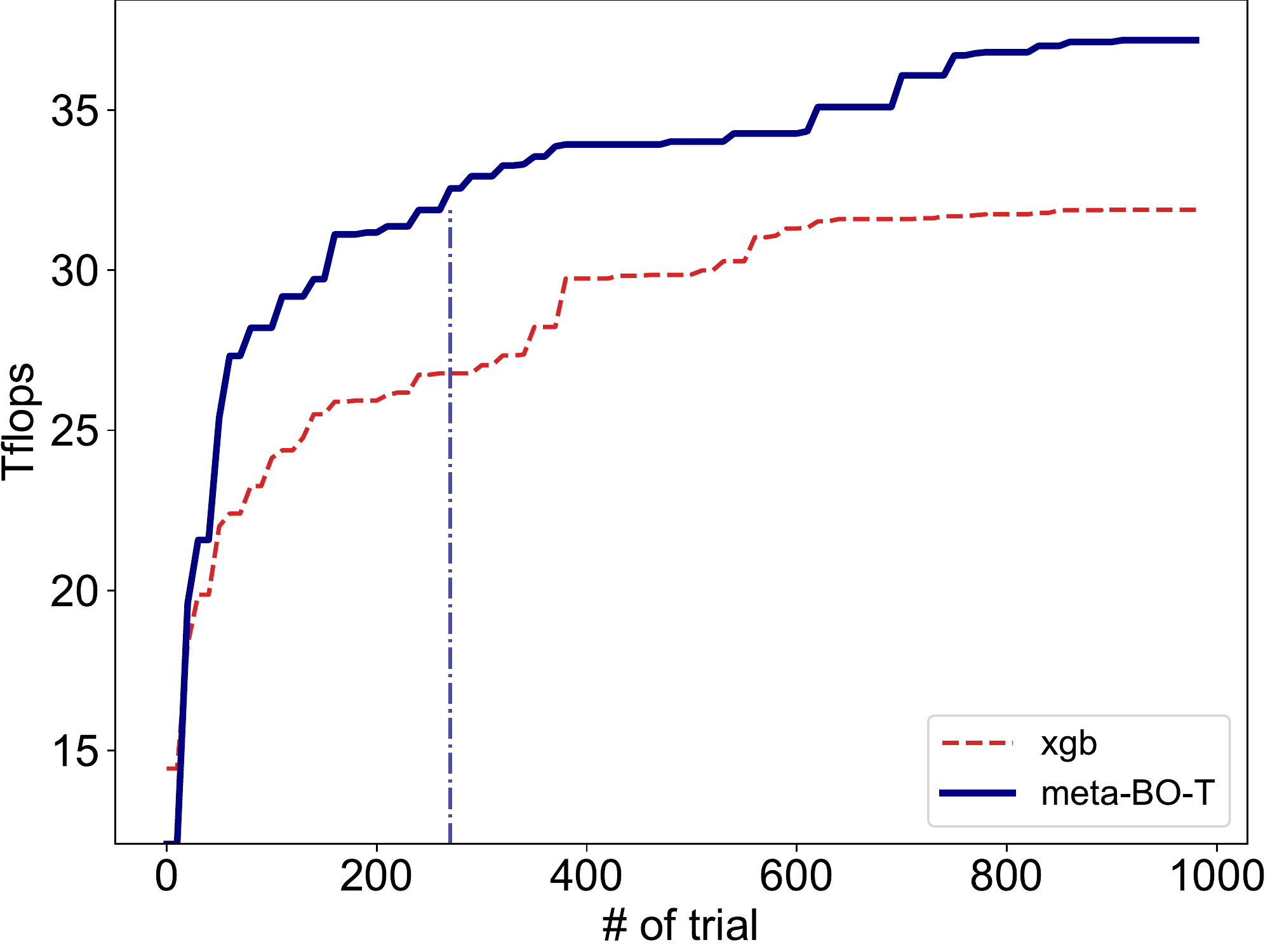}
\label{fig:amd:resnet-18}
}
\subfigure[Alexnet]{
\includegraphics[width=.47\textwidth]{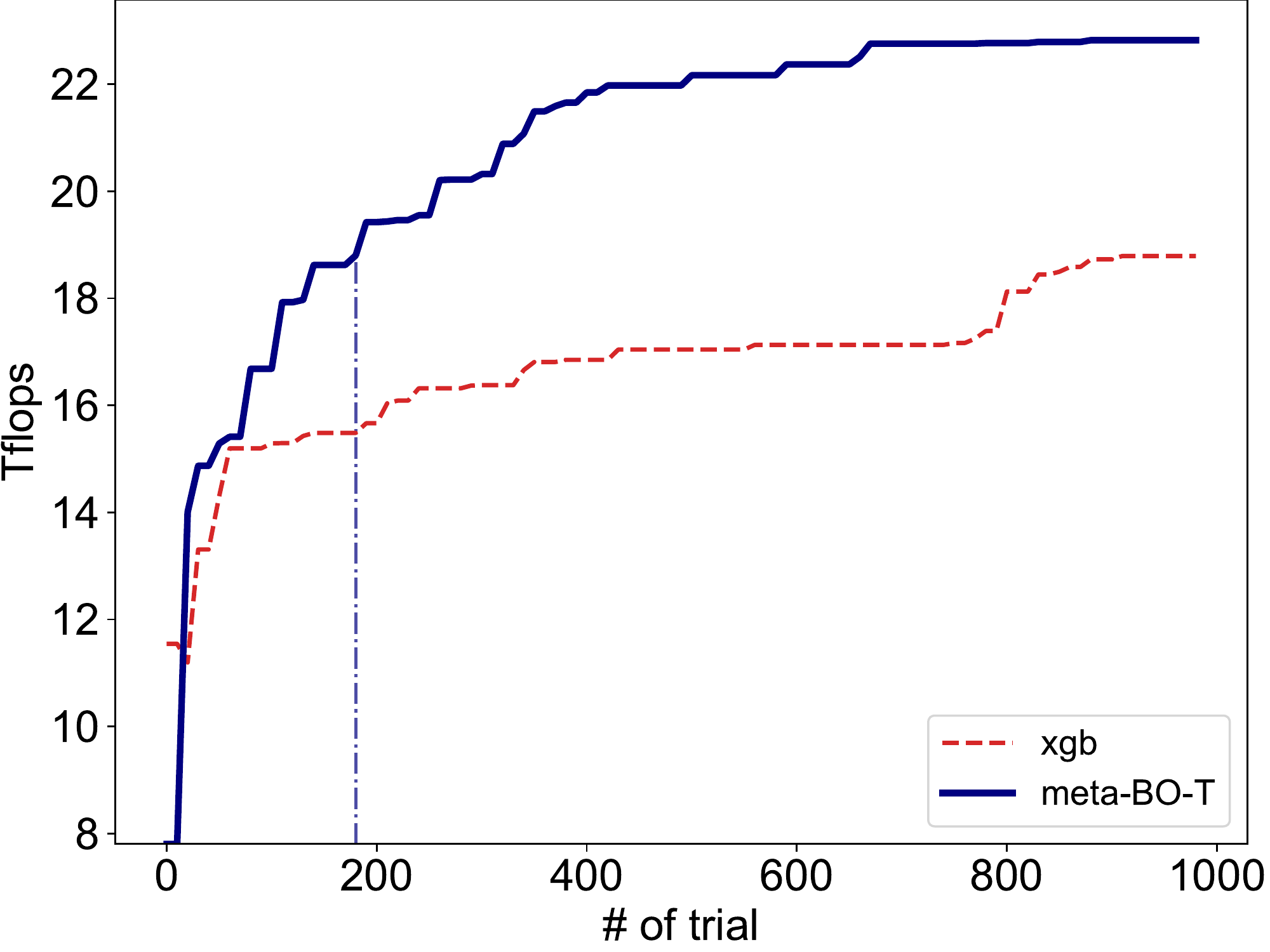}
\label{fig:amd:alexnet}
}
\caption{Output performance on AMD Vega 64 GPU}
\label{fig:amd:result}
\end{minipage}
\end{figure*}

\begin{figure}[!t]
\begin{center}
\includegraphics[width=.9\linewidth]{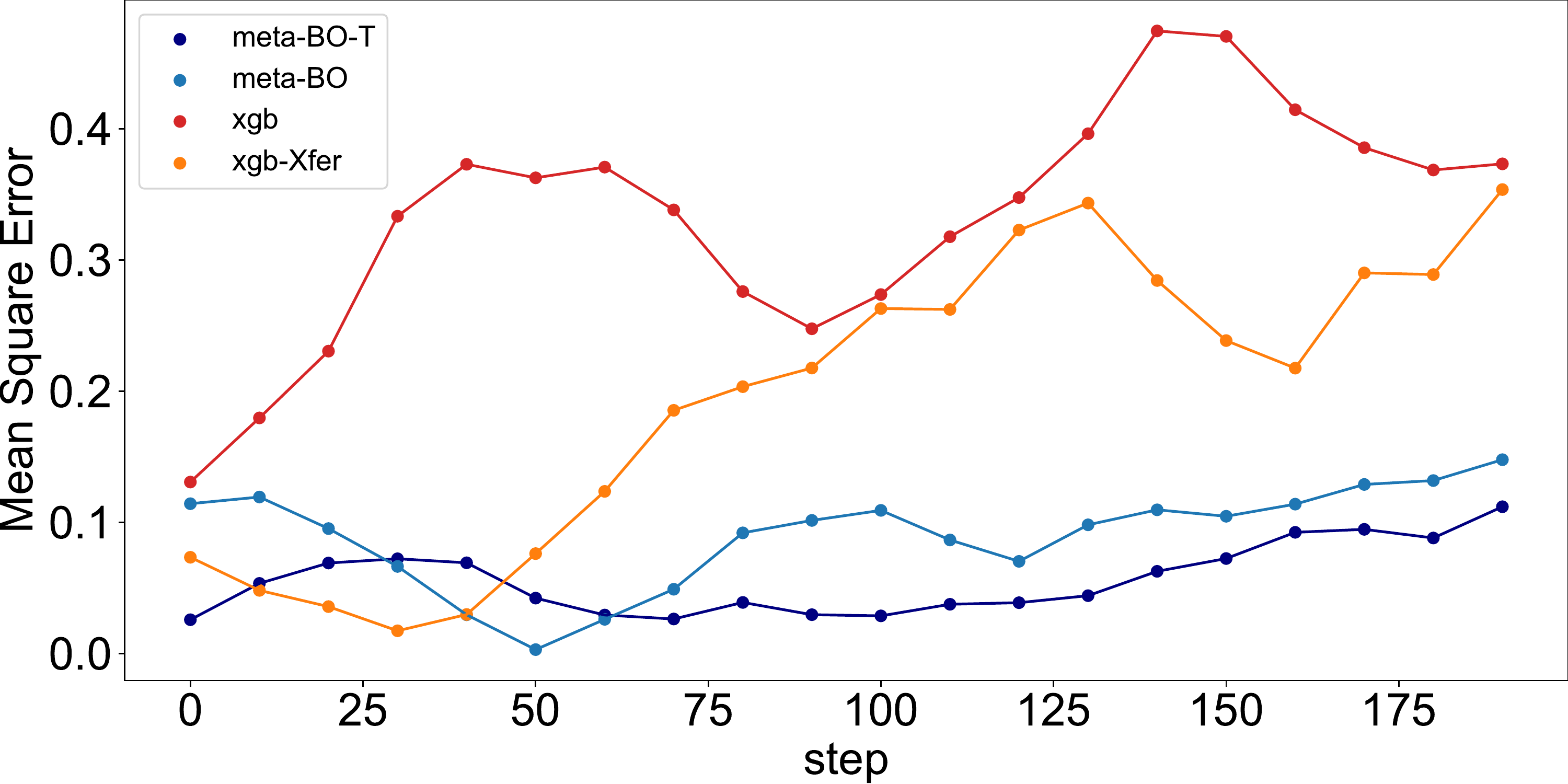}
\vspace{-0.1in}
\caption{Model accuracy in early iterations}
\label{fig:early_step}
\end{center}
\vspace{-0.15in}
\end{figure}

\begin{table}[]
\footnotesize
\centering
\begin{tabular}{p{1.3cm}|c|c|c|c}
\hline
 \textbf{}  & \textbf{xgb}  & \textbf{RL} & \textbf{ meta-BO } & \textbf{ meta-BO-T } \\ 
\hline
resnet-18 & 580.04 & 622.71 & \textbf{514.27}  & 576.21 \\ 
\hline
vgg-16 & 829.17 & 962.17 & \textbf{652.85}  & 727.76 \\ 
\hline
alexnet & 358.33 & 434.25 & \textbf{306.60}  & 341.61 \\ 
\hline
squeezenet & 966.78 & 1038.02 & \textbf{929.61}  & 980.15 \\
\hline
\end{tabular}
\vspace{-0.1in}
\caption{End-to-end optimization time (minutes)}
\label{fig:endtoend}
\vspace{-0.1in}
\end{table}

\subsection{Model Output Performance}

Figure~\ref{fig:ti:trial:result} shows the output performance for four CNN models in FLops, while Figure~\ref{fig:ti:bar:result} compares highest FLops achieved for the models normalized to {\it xgb}.
Overall, {\bf {\it meta-BO-T} provides the best performance for all configurations}, outperforming {\it xgb} by 8\% (up to 10\%) and {\it RL} by 13\% (up to 33\%) on average. {\it meta-BO} without super-graph augmentation also consistently improves {\it xgb} and {\it RL} by 5\% and 10\% on average. 
{\it meta-BO-T} consistently outperforming {\it meta-BO} by 4\% on average supports our assumption that super-graph augmentation helps with a quick adaptation of meta-learning models by increasing structural similarity in input data. 

You can see from the figure that both {\it meta-BO} and {\it meta-BO-T} adapt much faster to the compiled kernel than the others, i.e., move to points more likely to perform well and start candidate search from them, and reach the highest FLops predicted by {\it xgb} earlier between 200 to 650 iterations (dotted lines). As a result, MetaTune can search the space more effectively and widely with the given number of iteration locating better-performing optimization parameters. 

Both MetaTune and {\it xgb-Xfer} reuse previous experience for the current task. While {\it xgb-Xfer} slightly improves output performance against {\it xgb}, all MetaTune models show a more noticeable performance impact of the pre-trained model by outperforming {\it xgb-Xfer} by 4 to 7\% on average. 

MetaTune with RL-based search algorithm in~\cite{chameleon} instead of batch BO significantly outperforms {\it RL} by 7\% on average (up to 28\%), isolating the performance impact of the cost model. In this scenario, {\it meta-RL} produces better results than {\it meta-RL-T} for all evaluated models. We identified the source in adaptive sampler in~\cite{chameleon}, which performs much fewer hardware measurements, thus adaptation for {\it meta-RL-T} than {\it meta-RL} after filtering. We plan to investigate the issue further, but it is out of the scope of this paper. 

The evaluation results on AMD GPU as shown in Figure~\ref{fig:amd:alexnet} and~\ref{fig:amd:resnet-18} shows 18\% and 5\% performance improvement by {\it meta-BO-T} against {\it xgb}. This confirms that meta-learning based cost models are a portable solution whose optimization capability is not platform-dependent.

\subsection{Optimization Time}
Table~\ref{fig:endtoend} shows the end-to-end auto-tuning time for {\it xgb}, {\it RL}, {\it meta-BO}, and {\it meta-BO-T}. While {\it meta-BO} is the fastest while all four are in a close range (We could not reproduce the results for~\cite{chameleon}). {\it meta-BO-T} seems to incur more inference overheads with larger super-graph data than {\it meta-BO}, but is comparable or faster than {\it xgb} or {\it RL}.

\begin{table*}[!thb]
\centering

\begin{tabular}{c|c|c|c|c|c|c|c} 
\toprule
\textbf{network}  & \textbf{method}  & \textbf{MSE}  & \textbf{MSE(D)}  & \textbf{network}  & \textbf{method}  & \textbf{MSE}  & \textbf{MSE(D)}  \\ 
\hhline{========}
alexnet & xgb & 0.1329 & 0.5884 & resnet-18 & xgb & 0.1175 & 0.6318 \\
alexnet & xgb-Xfer & 0.1394 & 0.5289 & resnet-18 & xgb-Xfer & 0.1281 & 0.5666 \\
alexnet & meta-BO & 0.1342 & 0.5888 & resnet-18 & meta-BO & \textbf{0.1048 } & 0.5720 \\
alexnet & meta-BO-T & \textbf{0.1134 } & \textbf{0.4733 } & resnet-18 & meta-BO-T & 0.1061 & \textbf{0.5373 } \\ 
\hhline{========}
vgg-16 & xgb & 0.1223 & 0.5859 & squeezenet & xgb & \textbf{0.1225 } & 0.6058 \\
vgg-16 & xgb-Xfer & 0.1388 & 0.5415 & squeezenet & xgb-Xfer & 0.1268 & 0.5442 \\
vgg-16 & meta-BO & \textbf{0.1075 } & 0.5150 & squeezenet & meta-BO & 0.1288 & 0.4212 \\
vgg-16 & meta-BO-T & 0.1198 & \textbf{0.4754 } & squeezenet & meta-BO-T & 0.1485 & \textbf{0.4044 } \\ 
\hhline{========}
alexnet(X) & xgb & \textbf{0.1292 } & 0.4787 & resnet-18(X) & xgb & 0.1304 & 0.5600 \\
alexnet(X) & xgb-Xfer & 0.1384 & 0.4621 & resnet-18(X) & xgb-Xfer & 0.1335 & 0.5471 \\
alexnet(X) & meta-BO-T & 0.1337 & \textbf{0.3178 } & resnet-18(X) & meta-BO-T & \textbf{0.1177 } & \textbf{0.4319 } \\ 
\hhline{========}
alexnet(A) & xgb & 0.1018 & 0.7129 & resnet-18(A) & xgb & 0.1134 & 0.6264 \\
alexnet(A) & meta-BO-T & \textbf{0.082 } & \textbf{0.6150 } & resnet-18(A) & meta-BO-T & \textbf{0.0942 } & \textbf{0.6188 } \\
\hline
\end{tabular}

\caption{Model error for {\it xgb} and {\it xgb-Xfer}  vs. {\it meta-BO-T}. (X) for results on Titan XP, (A) on AMD GPU. }
\vspace{-0.1in}
\label{tab:eval2}
\end{table*}

\subsection{Model Accuracy} 

Table~\ref{tab:eval2} shows the prediction accuracy of the MetaTune cost model and the gradient-boosting based cost model in TVM through 1,000 iterations. 
We analyzed the inconsistency with MSE distribution and found out that infeasible or erroneous candidates randomly appear during exploration, and failing to predict performance for them unduly increases the model's MSE. Therefore, we introduced an additional accuracy metric than MSE; MSE(D) is a mean-square error of fine-tuning instances whose hardware-measured FLops are in the top 25\%.
This metric is useful in focusing on how effective the cost model is in identifying possibly best-performing candidates and guiding the search in the right direction. {\it meta-BO-T} shows the highest MSE(D) for all models on all platforms, which aligns with the output performance result.

We also investigated the model accuracy in early auto-tuning iterations in detail to see if MetaTune adapts more quickly than non-meta-learning models. From Figure~\ref{fig:early_step}, you can see that the MetaTune models start at a lower MSE and maintain high accuracy while {\it xgb} and {\it xgb-Xfer} may explore further away from ideal optimizations with more wrong predictions.

\subsection{Cross-Platform Meta-Learning}
We experimented with a pre-trained MetaTune model to see if it can adapt cross-platform as well, i.e., if meta-learned model parameters can universally work for performance distributions on which the model is not trained. Figure~\ref{fig:TITAN:resnet-18} and~\ref{fig:TITAN:alexnet} show the output performance (inference time) for \texttt{alexnet} and \texttt{resnet-18} on NVIDIA Titan XP GPU where {\it meta-BO-T} are pre-trained on NVIDIA 2080 Ti GPU. {\it meta-BO-T} provides 14\% and 6\% higher performance than {\it xgb} and {\it xgb-Xfer} showing that knowledge reuse across platforms is possible. Cross-platform meta-learning across more heterogeneous platforms is a part of our future work.

\section{Related Work}
\label{sec:related}

\subsection*{Adaptive exploration for automatic optimizations}

There has been active research in improving the exploration module in auto-tuning frameworks to reduce the total optimization time.
Chameleon~\cite{chameleon} addressed the issues of not selectively exploring feasible optimizations only ~\cite{autotvm} by proposing adaptive sampling and exploration algorithms using reinforcement learning
 ALT~\cite{alt} paid attention to the suboptimality caused by a premature cost model during the exploration process. It proposes a filtering mechanism for useless candidates that uses score prediction and active learning. 
AdaTune~\cite{adatune} addresses the exploration vs. exploitation problem that affects total optimization and time and output quality. It proposes surrogate modeling to solve the problem using uncertainty quantification and an evaluator that determines early stopping based on the coefficient variation. 

These are orthogonal to MetaTune focusing on the cost model and can be combined with MetaTune for further optimizations.

\subsection*{Auto-scheduling for tensor programs}

FlexTensor~\cite{10.1145/3373376.3378508} aims to generate dynamic schedules for tensor operations and perform automatic optimization with them on heterogeneous systems. Schedules are generated without code templates, focusing on the search space of each operation at a time. While this approach may locate more flexible and potentially efficient codes than using template-based schedules, it incurs much higher overheads with fine-grained searches. Our results show that the use of templates judiciously restricts the search space while not hurting the quality of the solution.
On a related note, Ansor~\cite{258858} also automatically generates optimization candidates hierarchically without template-based restrictions. There are no predefined knobs per operation, and a task scheduler learns the right level of hardware measurements as well. This approach requires a larger number of hardware measurements at each hierarchical level, ranging from 300 to 6,000 iterations.

\subsection*{Meta learning for fast adaptation}

Recent research used MAML to initialize model parameters for quick adaption in various applications.

\cite{winata2020learning} used MAML to adapt to the high variability and complex characteristics of accents for speech recognition.
T-NAS~\cite{tnas} leveraged MAML to handle multiple tasks in network architecture search, while
MAML is used to improve the accuracy of a super-resolution model for unseen images in ~\cite{mzsr}.

\section{Conclusion and Future Work}
\label{sec:conclusion}

In this paper, we presented a meta-learning based approach to improve the efficiency and reduce the overheads of the auto-tuning process by reusing prior experiences. Our experimental results showed that the resulting cost model, MetaTune, can identify high-performing optimization parameters for unseen convolution operations faster and more accurately than prior work, regardless of search algorithms used together. 
We believe that the meta-learning based cost model, or any machine-learning based model that can systematically leverage previously learned knowledge and experiences with similar programs, is a pivotal contribution to the auto-tuning solutions, addressing the high optimization overheads and allowing them to be a more widely-adopted compiler solution. 

Our future work includes extending MetaTune to support a wider range of tensor operations beyond convolution operations on different platforms. 
We believe that the meta-learning based approach can show even stronger and more consistent performance than related work with auto-tuning targets with varying static and dynamic characteristics. 
We also consider going out of the boundary of domain-specific compiler frameworks and looking more into solving more general optimization problems with MetaTune approach. 

\section*{Acknowledgement}
This work was supported by Institute for Information \& communications Technology Promotion (IITP) grant funded by the Korea government (MSIP) (No. 2019-0-01906, Artificial Intelligence Graduate School Program (POSTECH)) and the Super Computer Development Leading Program of the National Research Foundation of Korea (NRF) funded by the Korean government (Ministry of Science and ICT (MSIT)) (No. 2020M3H6A1084853).

\bibliographystyle{ACM-Reference-Format}
\bibliography{main}


\begin{thebibliography}{33}


\ifx \showCODEN    \undefined \def \showCODEN     #1{\unskip}     \fi
\ifx \showDOI      \undefined \def \showDOI       #1{#1}\fi
\ifx \showISBNx    \undefined \def \showISBNx     #1{\unskip}     \fi
\ifx \showISBNxiii \undefined \def \showISBNxiii  #1{\unskip}     \fi
\ifx \showISSN     \undefined \def \showISSN      #1{\unskip}     \fi
\ifx \showLCCN     \undefined \def \showLCCN      #1{\unskip}     \fi
\ifx \shownote     \undefined \def \shownote      #1{#1}          \fi
\ifx \showarticletitle \undefined \def \showarticletitle #1{#1}   \fi
\ifx \showURL      \undefined \def \showURL       {\relax}        \fi
\providecommand\bibfield[2]{#2}
\providecommand\bibinfo[2]{#2}
\providecommand\natexlab[1]{#1}
\providecommand\showeprint[2][]{arXiv:#2}

\bibitem[\protect\citeauthoryear{Abadi, Agarwal, Barham, Brevdo, Chen, Citro,
  Corrado, Davis, Dean, Devin, Ghemawat, Goodfellow, Harp, Irving, Isard, Jia,
  Jozefowicz, Kaiser, Kudlur, Levenberg, Man\'{e}, Monga, Moore, Murray, Olah,
  Schuster, Shlens, Steiner, Sutskever, Talwar, Tucker, Vanhoucke, Vasudevan,
  Vi\'{e}gas, Vinyals, Warden, Wattenberg, Wicke, Yu, and Zheng}{Abadi
  et~al\mbox{.}}{2015}]%
        {tensorflow}
\bibfield{author}{\bibinfo{person}{Mart\'{\i}n Abadi}, \bibinfo{person}{Ashish
  Agarwal}, \bibinfo{person}{Paul Barham}, \bibinfo{person}{Eugene Brevdo},
  \bibinfo{person}{Zhifeng Chen}, \bibinfo{person}{Craig Citro},
  \bibinfo{person}{Greg~S. Corrado}, \bibinfo{person}{Andy Davis},
  \bibinfo{person}{Jeffrey Dean}, \bibinfo{person}{Matthieu Devin},
  \bibinfo{person}{Sanjay Ghemawat}, \bibinfo{person}{Ian Goodfellow},
  \bibinfo{person}{Andrew Harp}, \bibinfo{person}{Geoffrey Irving},
  \bibinfo{person}{Michael Isard}, \bibinfo{person}{Yangqing Jia},
  \bibinfo{person}{Rafal Jozefowicz}, \bibinfo{person}{Lukasz Kaiser},
  \bibinfo{person}{Manjunath Kudlur}, \bibinfo{person}{Josh Levenberg},
  \bibinfo{person}{Dandelion Man\'{e}}, \bibinfo{person}{Rajat Monga},
  \bibinfo{person}{Sherry Moore}, \bibinfo{person}{Derek Murray},
  \bibinfo{person}{Chris Olah}, \bibinfo{person}{Mike Schuster},
  \bibinfo{person}{Jonathon Shlens}, \bibinfo{person}{Benoit Steiner},
  \bibinfo{person}{Ilya Sutskever}, \bibinfo{person}{Kunal Talwar},
  \bibinfo{person}{Paul Tucker}, \bibinfo{person}{Vincent Vanhoucke},
  \bibinfo{person}{Vijay Vasudevan}, \bibinfo{person}{Fernanda Vi\'{e}gas},
  \bibinfo{person}{Oriol Vinyals}, \bibinfo{person}{Pete Warden},
  \bibinfo{person}{Martin Wattenberg}, \bibinfo{person}{Martin Wicke},
  \bibinfo{person}{Yuan Yu}, {and} \bibinfo{person}{Xiaoqiang Zheng}.}
  \bibinfo{year}{2015}\natexlab{}.
\newblock \bibinfo{title}{{TensorFlow}: Large-Scale Machine Learning on
  Heterogeneous Systems}.
\newblock
\newblock
\urldef\tempurl%
\url{https://www.tensorflow.org/}
\showURL{%
\tempurl}
\newblock
\shownote{Software available from tensorflow.org.}


\bibitem[\protect\citeauthoryear{Ahn, Pilligundla, Yazdanbakhsh, and
  Esmaeilzadeh}{Ahn et~al\mbox{.}}{2020}]%
        {chameleon}
\bibfield{author}{\bibinfo{person}{Byung~Hoon Ahn}, \bibinfo{person}{Prannoy
  Pilligundla}, \bibinfo{person}{Amir Yazdanbakhsh}, {and}
  \bibinfo{person}{Hadi Esmaeilzadeh}.} \bibinfo{year}{2020}\natexlab{}.
\newblock \bibinfo{title}{Chameleon: Adaptive Code Optimization for Expedited
  Deep Neural Network Compilation}.
\newblock
\newblock
\showeprint[arxiv]{2001.08743}~[cs.LG]


\bibitem[\protect\citeauthoryear{Bertsimas, Tsitsiklis,
  et~al\mbox{.}}{Bertsimas et~al\mbox{.}}{1993}]%
        {sa}
\bibfield{author}{\bibinfo{person}{Dimitris Bertsimas}, \bibinfo{person}{John
  Tsitsiklis}, {et~al\mbox{.}}} \bibinfo{year}{1993}\natexlab{}.
\newblock \showarticletitle{Simulated annealing}.
\newblock \bibinfo{journal}{\emph{Statistical science}} \bibinfo{volume}{8},
  \bibinfo{number}{1} (\bibinfo{year}{1993}), \bibinfo{pages}{10--15}.
\newblock


\bibitem[\protect\citeauthoryear{Chen and Guestrin}{Chen and Guestrin}{2016}]%
        {xgb}
\bibfield{author}{\bibinfo{person}{Tianqi Chen} {and} \bibinfo{person}{Carlos
  Guestrin}.} \bibinfo{year}{2016}\natexlab{}.
\newblock \showarticletitle{XGBoost: A Scalable Tree Boosting System}. In
  \bibinfo{booktitle}{\emph{Proceedings of the 22nd ACM SIGKDD International
  Conference on Knowledge Discovery and Data Mining}} (San Francisco,
  California, USA) \emph{(\bibinfo{series}{KDD '16})}.
  \bibinfo{publisher}{Association for Computing Machinery},
  \bibinfo{address}{New York, NY, USA}, \bibinfo{pages}{785–794}.
\newblock
\showISBNx{9781450342322}
\urldef\tempurl%
\url{https://doi.org/10.1145/2939672.2939785}
\showDOI{\tempurl}


\bibitem[\protect\citeauthoryear{Chen, Li, Li, Lin, Wang, Wang, Xiao, Xu,
  Zhang, and Zhang}{Chen et~al\mbox{.}}{2015}]%
        {mxnet}
\bibfield{author}{\bibinfo{person}{Tianqi Chen}, \bibinfo{person}{Mu Li},
  \bibinfo{person}{Yutian Li}, \bibinfo{person}{Min Lin},
  \bibinfo{person}{Naiyan Wang}, \bibinfo{person}{Minjie Wang},
  \bibinfo{person}{Tianjun Xiao}, \bibinfo{person}{Bing Xu},
  \bibinfo{person}{Chiyuan Zhang}, {and} \bibinfo{person}{Zheng Zhang}.}
  \bibinfo{year}{2015}\natexlab{}.
\newblock \showarticletitle{MXNet: A Flexible and Efficient Machine Learning
  Library for Heterogeneous Distributed Systems.}
\newblock \bibinfo{journal}{\emph{CoRR}}  \bibinfo{volume}{abs/1512.01274}
  (\bibinfo{year}{2015}).
\newblock
\urldef\tempurl%
\url{http://dblp.uni-trier.de/db/journals/corr/corr1512.html#ChenLLLWWXXZZ15}
\showURL{%
\tempurl}


\bibitem[\protect\citeauthoryear{Chen, Moreau, Jiang, Zheng, Yan, Cowan, Shen,
  Wang, Hu, Ceze, Guestrin, and Krishnamurthy}{Chen et~al\mbox{.}}{2018a}]%
        {tvm}
\bibfield{author}{\bibinfo{person}{Tianqi Chen}, \bibinfo{person}{Thierry
  Moreau}, \bibinfo{person}{Ziheng Jiang}, \bibinfo{person}{Lianmin Zheng},
  \bibinfo{person}{Eddie Yan}, \bibinfo{person}{Meghan Cowan},
  \bibinfo{person}{Haichen Shen}, \bibinfo{person}{Leyuan Wang},
  \bibinfo{person}{Yuwei Hu}, \bibinfo{person}{Luis Ceze},
  \bibinfo{person}{Carlos Guestrin}, {and} \bibinfo{person}{Arvind
  Krishnamurthy}.} \bibinfo{year}{2018}\natexlab{a}.
\newblock \showarticletitle{TVM: An Automated End-to-End Optimizing Compiler
  for Deep Learning}. In \bibinfo{booktitle}{\emph{Proceedings of the 13th
  USENIX Conference on Operating Systems Design and Implementation}} (Carlsbad,
  CA, USA) \emph{(\bibinfo{series}{OSDI'18})}. \bibinfo{publisher}{USENIX
  Association}, \bibinfo{address}{USA}, \bibinfo{pages}{579–594}.
\newblock
\showISBNx{9781931971478}


\bibitem[\protect\citeauthoryear{Chen, Zheng, Yan, Jiang, Moreau, Ceze,
  Guestrin, and Krishnamurthy}{Chen et~al\mbox{.}}{2018b}]%
        {autotvm}
\bibfield{author}{\bibinfo{person}{Tianqi Chen}, \bibinfo{person}{Lianmin
  Zheng}, \bibinfo{person}{Eddie Yan}, \bibinfo{person}{Ziheng Jiang},
  \bibinfo{person}{Thierry Moreau}, \bibinfo{person}{Luis Ceze},
  \bibinfo{person}{Carlos Guestrin}, {and} \bibinfo{person}{Arvind
  Krishnamurthy}.} \bibinfo{year}{2018}\natexlab{b}.
\newblock \showarticletitle{Learning to Optimize Tensor Programs}. In
  \bibinfo{booktitle}{\emph{Proceedings of the 32nd International Conference on
  Neural Information Processing Systems}} (Montr\'{e}al, Canada)
  \emph{(\bibinfo{series}{NIPS'18})}. \bibinfo{publisher}{Curran Associates
  Inc.}, \bibinfo{address}{Red Hook, NY, USA}, \bibinfo{pages}{3393–3404}.
\newblock


\bibitem[\protect\citeauthoryear{{Cummins}, {Petoumenos}, {Wang}, and
  {Leather}}{{Cummins} et~al\mbox{.}}{2017}]%
        {Cummins2017}
\bibfield{author}{\bibinfo{person}{C. {Cummins}}, \bibinfo{person}{P.
  {Petoumenos}}, \bibinfo{person}{Z. {Wang}}, {and} \bibinfo{person}{H.
  {Leather}}.} \bibinfo{year}{2017}\natexlab{}.
\newblock \showarticletitle{End-to-End Deep Learning of Optimization
  Heuristics}. In \bibinfo{booktitle}{\emph{2017 26th International Conference
  on Parallel Architectures and Compilation Techniques (PACT)}}.
  \bibinfo{pages}{219--232}.
\newblock
\urldef\tempurl%
\url{https://doi.org/10.1109/PACT.2017.24}
\showDOI{\tempurl}


\bibitem[\protect\citeauthoryear{Desautels, Krause, and Burdick}{Desautels
  et~al\mbox{.}}{2014}]%
        {Desautels:JMLR:2014}
\bibfield{author}{\bibinfo{person}{Thomas Desautels}, \bibinfo{person}{Andreas
  Krause}, {and} \bibinfo{person}{Joel~W. Burdick}.}
  \bibinfo{year}{2014}\natexlab{}.
\newblock \showarticletitle{Parallelizing Exploration-Exploitation Tradeoffs in
  Gaussian Process Bandit Optimization}.
\newblock \bibinfo{journal}{\emph{J. Mach. Learn. Res.}} \bibinfo{volume}{15},
  \bibinfo{number}{1} (\bibinfo{date}{Jan.} \bibinfo{year}{2014}),
  \bibinfo{pages}{3873–3923}.
\newblock
\showISSN{1532-4435}


\bibitem[\protect\citeauthoryear{Dhakal, Cho, Kulkarni, Ramakrishnan, and
  Sharma}{Dhakal et~al\mbox{.}}{2020}]%
        {dhakal2020spatial}
\bibfield{author}{\bibinfo{person}{Aditya Dhakal}, \bibinfo{person}{Junguk
  Cho}, \bibinfo{person}{Sameer~G. Kulkarni}, \bibinfo{person}{K.~K.
  Ramakrishnan}, {and} \bibinfo{person}{Puneet Sharma}.}
  \bibinfo{year}{2020}\natexlab{}.
\newblock \bibinfo{title}{Spatial Sharing of GPU for Autotuning DNN models}.
\newblock
\newblock
\showeprint[arxiv]{2008.03602}~[cs.NE]


\bibitem[\protect\citeauthoryear{Ding, Ansel, Veeramachaneni, Shen, O’Reilly,
  and Amarasinghe}{Ding et~al\mbox{.}}{2015a}]%
        {Autotuning1}
\bibfield{author}{\bibinfo{person}{Yufei Ding}, \bibinfo{person}{Jason Ansel},
  \bibinfo{person}{Kalyan Veeramachaneni}, \bibinfo{person}{Xipeng Shen},
  \bibinfo{person}{Una-May O’Reilly}, {and} \bibinfo{person}{Saman
  Amarasinghe}.} \bibinfo{year}{2015}\natexlab{a}.
\newblock \showarticletitle{Autotuning Algorithmic Choice for Input
  Sensitivity}. In \bibinfo{booktitle}{\emph{Proceedings of the 36th ACM
  SIGPLAN Conference on Programming Language Design and Implementation}}
  (Portland, OR, USA) \emph{(\bibinfo{series}{PLDI '15})}.
  \bibinfo{publisher}{Association for Computing Machinery},
  \bibinfo{address}{New York, NY, USA}, \bibinfo{pages}{379–390}.
\newblock
\showISBNx{9781450334686}
\urldef\tempurl%
\url{https://doi.org/10.1145/2737924.2737969}
\showDOI{\tempurl}


\bibitem[\protect\citeauthoryear{Ding, Ansel, Veeramachaneni, Shen, O’Reilly,
  and Amarasinghe}{Ding et~al\mbox{.}}{2015b}]%
        {Autotuning2}
\bibfield{author}{\bibinfo{person}{Yufei Ding}, \bibinfo{person}{Jason Ansel},
  \bibinfo{person}{Kalyan Veeramachaneni}, \bibinfo{person}{Xipeng Shen},
  \bibinfo{person}{Una-May O’Reilly}, {and} \bibinfo{person}{Saman
  Amarasinghe}.} \bibinfo{year}{2015}\natexlab{b}.
\newblock \showarticletitle{Autotuning Algorithmic Choice for Input
  Sensitivity}.
\newblock \bibinfo{journal}{\emph{SIGPLAN Not.}} \bibinfo{volume}{50},
  \bibinfo{number}{6} (\bibinfo{date}{June} \bibinfo{year}{2015}),
  \bibinfo{pages}{379–390}.
\newblock
\showISSN{0362-1340}
\urldef\tempurl%
\url{https://doi.org/10.1145/2813885.2737969}
\showDOI{\tempurl}


\bibitem[\protect\citeauthoryear{Finn, Abbeel, and Levine}{Finn
  et~al\mbox{.}}{2017}]%
        {maml}
\bibfield{author}{\bibinfo{person}{Chelsea Finn}, \bibinfo{person}{Pieter
  Abbeel}, {and} \bibinfo{person}{Sergey Levine}.}
  \bibinfo{year}{2017}\natexlab{}.
\newblock \bibinfo{title}{Model-Agnostic Meta-Learning for Fast Adaptation of
  Deep Networks}.
\newblock
\newblock
\showeprint[arxiv]{1703.03400}~[cs.LG]


\bibitem[\protect\citeauthoryear{Fursin, Kashnikov, Memon, Chamski, Temam,
  Namolaru, Yom-Tov, Mendelson, Zaks, Courtois, Bodin, Barnard, Ashton,
  Bonilla, Thomson, Williams, and O'Boyle}{Fursin et~al\mbox{.}}{2011}]%
        {milepost}
\bibfield{author}{\bibinfo{person}{Grigori Fursin}, \bibinfo{person}{Yuriy
  Kashnikov}, \bibinfo{person}{Abdul~Wahid Memon}, \bibinfo{person}{Zbigniew
  Chamski}, \bibinfo{person}{Olivier Temam}, \bibinfo{person}{Mircea Namolaru},
  \bibinfo{person}{Elad Yom-Tov}, \bibinfo{person}{Bilha Mendelson},
  \bibinfo{person}{Ayal Zaks}, \bibinfo{person}{Eric Courtois},
  \bibinfo{person}{François Bodin}, \bibinfo{person}{Phil Barnard},
  \bibinfo{person}{Elton Ashton}, \bibinfo{person}{Edwin~V. Bonilla},
  \bibinfo{person}{John Thomson}, \bibinfo{person}{Christopher K.~I. Williams},
  {and} \bibinfo{person}{Michael F.~P. O'Boyle}.}
  \bibinfo{year}{2011}\natexlab{}.
\newblock \showarticletitle{Milepost GCC: Machine Learning Enabled Self-tuning
  Compiler.}
\newblock \bibinfo{journal}{\emph{Int. J. Parallel Program.}}
  \bibinfo{volume}{39}, \bibinfo{number}{3} (\bibinfo{year}{2011}),
  \bibinfo{pages}{296--327}.
\newblock
\urldef\tempurl%
\url{http://dblp.uni-trier.de/db/journals/ijpp/ijpp39.html#FursinKMCTNYMZCa11}
\showURL{%
\tempurl}


\bibitem[\protect\citeauthoryear{Kaufman, Phothilimthana, Zhou, and
  Burrows}{Kaufman et~al\mbox{.}}{2020}]%
        {kaufman2020learned}
\bibfield{author}{\bibinfo{person}{Samuel~J. Kaufman},
  \bibinfo{person}{Phitchaya~Mangpo Phothilimthana}, \bibinfo{person}{Yanqi
  Zhou}, {and} \bibinfo{person}{Mike Burrows}.}
  \bibinfo{year}{2020}\natexlab{}.
\newblock \bibinfo{title}{A Learned Performance Model for the Tensor Processing
  Unit}.
\newblock
\newblock
\showeprint[arxiv]{2008.01040}~[cs.PF]


\bibitem[\protect\citeauthoryear{Kipf and Welling}{Kipf and Welling}{2017}]%
        {gcn}
\bibfield{author}{\bibinfo{person}{Thomas~N. Kipf} {and} \bibinfo{person}{Max
  Welling}.} \bibinfo{year}{2017}\natexlab{}.
\newblock \showarticletitle{{Semi-Supervised Classification with Graph
  Convolutional Networks}}. In \bibinfo{booktitle}{\emph{Proceedings of the 5th
  International Conference on Learning Representations}} (Palais des
  Congr{\`e}s Neptune, Toulon, France) \emph{(\bibinfo{series}{ICLR '17})}.
\newblock
\urldef\tempurl%
\url{https://openreview.net/forum?id=SJU4ayYgl}
\showURL{%
\tempurl}


\bibitem[\protect\citeauthoryear{Koch}{Koch}{2015}]%
        {simaese}
\bibfield{author}{\bibinfo{person}{Gregory Koch}.}
  \bibinfo{year}{2015}\natexlab{}.
\newblock \showarticletitle{Siamese neural networks for one-shot image
  recognition}.
\newblock


\bibitem[\protect\citeauthoryear{Leary and Wang}{Leary and Wang}{2017}]%
        {xla}
\bibfield{author}{\bibinfo{person}{Chris Leary} {and} \bibinfo{person}{Todd
  Wang}.} \bibinfo{year}{2017}\natexlab{}.
\newblock \showarticletitle{XLA: TensorFlow, compiled}.
\newblock \bibinfo{journal}{\emph{TensorFlow Dev Summit}}
  (\bibinfo{year}{2017}).
\newblock


\bibitem[\protect\citeauthoryear{Li, Zhang, Wang, and Li}{Li
  et~al\mbox{.}}{2020}]%
        {adatune}
\bibfield{author}{\bibinfo{person}{Menghao Li}, \bibinfo{person}{Minjia Zhang},
  \bibinfo{person}{Chi Wang}, {and} \bibinfo{person}{Mingqin Li}.}
  \bibinfo{year}{2020}\natexlab{}.
\newblock \showarticletitle{AdaTune: Adaptive Tensor Program Compilation Made
  Efficient}. In \bibinfo{booktitle}{\emph{34th Conference on Neural
  Information Processing Systems (NeurIPS 2020)}}.
\newblock
\urldef\tempurl%
\url{https://www.microsoft.com/en-us/research/publication/adatune-adaptive-tensor-program-compilation-made-efficient/}
\showURL{%
\tempurl}


\bibitem[\protect\citeauthoryear{Lian, Yin~Zheng, Lu, Lin, Zhao, Huang, and
  Gao.}{Lian et~al\mbox{.}}{2020}]%
        {tnas}
\bibfield{author}{\bibinfo{person}{Dongze Lian}, \bibinfo{person}{Yintao~Xu
  Yin~Zheng}, \bibinfo{person}{Yanxiong Lu}, \bibinfo{person}{Leyu Lin},
  \bibinfo{person}{Peilin Zhao}, \bibinfo{person}{Junzhou Huang}, {and}
  \bibinfo{person}{Shenghua Gao.}} \bibinfo{year}{2020}\natexlab{}.
\newblock \showarticletitle{Towards Fast Adaptation of Neural Architectures
  with Meta Learning.}
\newblock \bibinfo{journal}{\emph{ICLR}} (\bibinfo{year}{2020}).
\newblock


\bibitem[\protect\citeauthoryear{Paszke, Gross, Massa, Lerer, Bradbury, Chanan,
  Killeen, Lin, Gimelshein, Antiga, Desmaison, Kopf, Yang, DeVito, Raison,
  Tejani, Chilamkurthy, Steiner, Fang, Bai, and Chintala}{Paszke
  et~al\mbox{.}}{2019}]%
        {pytorch}
\bibfield{author}{\bibinfo{person}{Adam Paszke}, \bibinfo{person}{Sam Gross},
  \bibinfo{person}{Francisco Massa}, \bibinfo{person}{Adam Lerer},
  \bibinfo{person}{James Bradbury}, \bibinfo{person}{Gregory Chanan},
  \bibinfo{person}{Trevor Killeen}, \bibinfo{person}{Zeming Lin},
  \bibinfo{person}{Natalia Gimelshein}, \bibinfo{person}{Luca Antiga},
  \bibinfo{person}{Alban Desmaison}, \bibinfo{person}{Andreas Kopf},
  \bibinfo{person}{Edward Yang}, \bibinfo{person}{Zachary DeVito},
  \bibinfo{person}{Martin Raison}, \bibinfo{person}{Alykhan Tejani},
  \bibinfo{person}{Sasank Chilamkurthy}, \bibinfo{person}{Benoit Steiner},
  \bibinfo{person}{Lu Fang}, \bibinfo{person}{Junjie Bai}, {and}
  \bibinfo{person}{Soumith Chintala}.} \bibinfo{year}{2019}\natexlab{}.
\newblock \showarticletitle{PyTorch: An Imperative Style, High-Performance Deep
  Learning Library}.
\newblock In \bibinfo{booktitle}{\emph{Advances in Neural Information
  Processing Systems 32}}, \bibfield{editor}{\bibinfo{person}{H.~Wallach},
  \bibinfo{person}{H.~Larochelle}, \bibinfo{person}{A.~Beygelzimer},
  \bibinfo{person}{F.~d\textquotesingle Alch\'{e}-Buc},
  \bibinfo{person}{E.~Fox}, {and} \bibinfo{person}{R.~Garnett}} (Eds.).
  \bibinfo{publisher}{Curran Associates, Inc.}, \bibinfo{pages}{8024--8035}.
\newblock
\urldef\tempurl%
\url{http://papers.neurips.cc/paper/9015-pytorch-an-imperative-style-high-performance-deep-learning-library.pdf}
\showURL{%
\tempurl}


\bibitem[\protect\citeauthoryear{Ragan-Kelley, Barnes, Adams, Paris, Durand,
  and Amarasinghe}{Ragan-Kelley et~al\mbox{.}}{2013}]%
        {halide}
\bibfield{author}{\bibinfo{person}{Jonathan Ragan-Kelley},
  \bibinfo{person}{Connelly Barnes}, \bibinfo{person}{Andrew Adams},
  \bibinfo{person}{Sylvain Paris}, \bibinfo{person}{Fr\'{e}do Durand}, {and}
  \bibinfo{person}{Saman Amarasinghe}.} \bibinfo{year}{2013}\natexlab{}.
\newblock \showarticletitle{Halide: A Language and Compiler for Optimizing
  Parallelism, Locality, and Recomputation in Image Processing Pipelines}. In
  \bibinfo{booktitle}{\emph{Proceedings of the 34th ACM SIGPLAN Conference on
  Programming Language Design and Implementation}} (Seattle, Washington, USA)
  \emph{(\bibinfo{series}{PLDI '13})}. \bibinfo{publisher}{Association for
  Computing Machinery}, \bibinfo{address}{New York, NY, USA},
  \bibinfo{pages}{519–530}.
\newblock
\showISBNx{9781450320146}
\urldef\tempurl%
\url{https://doi.org/10.1145/2491956.2462176}
\showDOI{\tempurl}


\bibitem[\protect\citeauthoryear{Rotem, Fix, Abdulrasool, Catron, Deng,
  Dzhabarov, Gibson, Hegeman, Lele, Levenstein, Montgomery, Maher, Nadathur,
  Olesen, Park, Rakhov, Smelyanskiy, and Wang}{Rotem et~al\mbox{.}}{2019}]%
        {glow}
\bibfield{author}{\bibinfo{person}{Nadav Rotem}, \bibinfo{person}{Jordan Fix},
  \bibinfo{person}{Saleem Abdulrasool}, \bibinfo{person}{Garret Catron},
  \bibinfo{person}{Summer Deng}, \bibinfo{person}{Roman Dzhabarov},
  \bibinfo{person}{Nick Gibson}, \bibinfo{person}{James Hegeman},
  \bibinfo{person}{Meghan Lele}, \bibinfo{person}{Roman Levenstein},
  \bibinfo{person}{Jack Montgomery}, \bibinfo{person}{Bert Maher},
  \bibinfo{person}{Satish Nadathur}, \bibinfo{person}{Jakob Olesen},
  \bibinfo{person}{Jongsoo Park}, \bibinfo{person}{Artem Rakhov},
  \bibinfo{person}{Misha Smelyanskiy}, {and} \bibinfo{person}{Man Wang}.}
  \bibinfo{year}{2019}\natexlab{}.
\newblock \bibinfo{title}{Glow: Graph Lowering Compiler Techniques for Neural
  Networks}.
\newblock
\newblock
\showeprint[arxiv]{1805.00907}~[cs.PL]


\bibitem[\protect\citeauthoryear{Soh, Cho, and Cho}{Soh et~al\mbox{.}}{2020}]%
        {mzsr}
\bibfield{author}{\bibinfo{person}{Jae~Woong Soh}, \bibinfo{person}{Sunwoo
  Cho}, {and} \bibinfo{person}{Nam~Ik Cho}.} \bibinfo{year}{2020}\natexlab{}.
\newblock \showarticletitle{Meta-Transfer Learning for Zero-Shot
  Super-Resolution}. In \bibinfo{booktitle}{\emph{Proceedings of the IEEE/CVF
  Conference on Computer Vision and Pattern Recognition}}.
  \bibinfo{pages}{3516--3525}.
\newblock


\bibitem[\protect\citeauthoryear{{Sung}, {Chen}, {Eichenberger}, and
  {O'Brien}}{{Sung} et~al\mbox{.}}{2019}]%
        {cogr}
\bibfield{author}{\bibinfo{person}{H. {Sung}}, \bibinfo{person}{T. {Chen}},
  \bibinfo{person}{A. {Eichenberger}}, {and} \bibinfo{person}{K.~K.
  {O'Brien}}.} \bibinfo{year}{2019}\natexlab{}.
\newblock \showarticletitle{POSTER: CogR: Exploiting Program Structures for
  Machine-Learning Based Runtime Solutions}. In \bibinfo{booktitle}{\emph{2019
  28th International Conference on Parallel Architectures and Compilation
  Techniques (PACT)}}. \bibinfo{pages}{485--486}.
\newblock
\urldef\tempurl%
\url{https://doi.org/10.1109/PACT.2019.00057}
\showDOI{\tempurl}


\bibitem[\protect\citeauthoryear{Tomczak, Lepert, and Wiggers}{Tomczak
  et~al\mbox{.}}{2019}]%
        {qast}
\bibfield{author}{\bibinfo{person}{Jakub~M. Tomczak}, \bibinfo{person}{Romain
  Lepert}, {and} \bibinfo{person}{Auke Wiggers}.}
  \bibinfo{year}{2019}\natexlab{}.
\newblock \bibinfo{title}{Simulating Execution Time of Tensor Programs using
  Graph Neural Networks}.
\newblock
\newblock
\showeprint{arXiv:1904.11876}


\bibitem[\protect\citeauthoryear{Vasilache, Zinenko, Theodoridis, Goyal,
  DeVito, Moses, Verdoolaege, Adams, and Cohen}{Vasilache
  et~al\mbox{.}}{2018}]%
        {TensorComprehensions}
\bibfield{author}{\bibinfo{person}{Nicolas Vasilache},
  \bibinfo{person}{Oleksandr Zinenko}, \bibinfo{person}{Theodoros Theodoridis},
  \bibinfo{person}{Priya Goyal}, \bibinfo{person}{Zachary DeVito},
  \bibinfo{person}{William~S. Moses}, \bibinfo{person}{Sven Verdoolaege},
  \bibinfo{person}{Andrew Adams}, {and} \bibinfo{person}{Albert Cohen}.}
  \bibinfo{year}{2018}\natexlab{}.
\newblock \bibinfo{title}{Tensor Comprehensions: Framework-Agnostic
  High-Performance Machine Learning Abstractions}.
\newblock
\newblock
\showeprint[arxiv]{1802.04730}~[cs.PL]


\bibitem[\protect\citeauthoryear{Wilson, Moriconi, Hutter, and
  Deisenroth}{Wilson et~al\mbox{.}}{2017}]%
        {qucb}
\bibfield{author}{\bibinfo{person}{James~T. Wilson}, \bibinfo{person}{Riccardo
  Moriconi}, \bibinfo{person}{Frank Hutter}, {and} \bibinfo{person}{Marc~Peter
  Deisenroth}.} \bibinfo{year}{2017}\natexlab{}.
\newblock \bibinfo{title}{The reparameterization trick for acquisition
  functions}.
\newblock
\newblock
\showeprint[arxiv]{1712.00424}~[stat.ML]


\bibitem[\protect\citeauthoryear{Winata, Cahyawijaya, Liu, Lin, Madotto, Xu,
  and Fung}{Winata et~al\mbox{.}}{2020}]%
        {winata2020learning}
\bibfield{author}{\bibinfo{person}{Genta~Indra Winata}, \bibinfo{person}{Samuel
  Cahyawijaya}, \bibinfo{person}{Zihan Liu}, \bibinfo{person}{Zhaojiang Lin},
  \bibinfo{person}{Andrea Madotto}, \bibinfo{person}{Peng Xu}, {and}
  \bibinfo{person}{Pascale Fung}.} \bibinfo{year}{2020}\natexlab{}.
\newblock \bibinfo{title}{Learning Fast Adaptation on Cross-Accented Speech
  Recognition}.
\newblock
\newblock
\showeprint[arxiv]{2003.01901}~[eess.AS]


\bibitem[\protect\citeauthoryear{Wu and Frazier}{Wu and Frazier}{2016}]%
        {pbo}
\bibfield{author}{\bibinfo{person}{Jian Wu} {and} \bibinfo{person}{Peter~I.
  Frazier}.} \bibinfo{year}{2016}\natexlab{}.
\newblock \showarticletitle{The Parallel Knowledge Gradient Method for Batch
  Bayesian Optimization}. In \bibinfo{booktitle}{\emph{Proceedings of the 30th
  International Conference on Neural Information Processing Systems}}
  (Barcelona, Spain) \emph{(\bibinfo{series}{NIPS'16})}.
  \bibinfo{publisher}{Curran Associates Inc.}, \bibinfo{address}{Red Hook, NY,
  USA}, \bibinfo{pages}{3134–3142}.
\newblock
\showISBNx{9781510838819}


\bibitem[\protect\citeauthoryear{{Zeng}, {Zhi}, {Du}, {Guo}, {Sun}, and
  {Chen}}{{Zeng} et~al\mbox{.}}{2020}]%
        {alt}
\bibfield{author}{\bibinfo{person}{X. {Zeng}}, \bibinfo{person}{T. {Zhi}},
  \bibinfo{person}{Z. {Du}}, \bibinfo{person}{Q. {Guo}}, \bibinfo{person}{N.
  {Sun}}, {and} \bibinfo{person}{Y. {Chen}}.} \bibinfo{year}{2020}\natexlab{}.
\newblock \showarticletitle{ALT: Optimizing Tensor Compilation in Deep Learning
  Compilers with Active Learning}. In \bibinfo{booktitle}{\emph{2020 IEEE 38th
  International Conference on Computer Design (ICCD)}}.
  \bibinfo{pages}{623--630}.
\newblock
\urldef\tempurl%
\url{https://doi.org/10.1109/ICCD50377.2020.00108}
\showDOI{\tempurl}


\bibitem[\protect\citeauthoryear{Zheng, Jia, Sun, Wu, Yu, Haj-Ali, Wang, Yang,
  Zhuo, Sen, Gonzalez, and Stoica}{Zheng et~al\mbox{.}}{2020a}]%
        {258858}
\bibfield{author}{\bibinfo{person}{Lianmin Zheng}, \bibinfo{person}{Chengfan
  Jia}, \bibinfo{person}{Minmin Sun}, \bibinfo{person}{Zhao Wu},
  \bibinfo{person}{Cody~Hao Yu}, \bibinfo{person}{Ameer Haj-Ali},
  \bibinfo{person}{Yida Wang}, \bibinfo{person}{Jun Yang},
  \bibinfo{person}{Danyang Zhuo}, \bibinfo{person}{Koushik Sen},
  \bibinfo{person}{Joseph~E. Gonzalez}, {and} \bibinfo{person}{Ion Stoica}.}
  \bibinfo{year}{2020}\natexlab{a}.
\newblock \showarticletitle{Ansor: Generating High-Performance Tensor Programs
  for Deep Learning}. In \bibinfo{booktitle}{\emph{14th {USENIX} Symposium on
  Operating Systems Design and Implementation ({OSDI} 20)}}.
  \bibinfo{publisher}{{USENIX} Association}, \bibinfo{pages}{863--879}.
\newblock
\showISBNx{978-1-939133-19-9}
\urldef\tempurl%
\url{https://www.usenix.org/conference/osdi20/presentation/zheng}
\showURL{%
\tempurl}


\bibitem[\protect\citeauthoryear{Zheng, Liang, Wang, Chen, and Sheng}{Zheng
  et~al\mbox{.}}{2020b}]%
        {10.1145/3373376.3378508}
\bibfield{author}{\bibinfo{person}{Size Zheng}, \bibinfo{person}{Yun Liang},
  \bibinfo{person}{Shuo Wang}, \bibinfo{person}{Renze Chen}, {and}
  \bibinfo{person}{Kaiwen Sheng}.} \bibinfo{year}{2020}\natexlab{b}.
\newblock \showarticletitle{FlexTensor: An Automatic Schedule Exploration and
  Optimization Framework for Tensor Computation on Heterogeneous System}. In
  \bibinfo{booktitle}{\emph{Proceedings of the Twenty-Fifth International
  Conference on Architectural Support for Programming Languages and Operating
  Systems}} (Lausanne, Switzerland) \emph{(\bibinfo{series}{ASPLOS '20})}.
  \bibinfo{publisher}{Association for Computing Machinery},
  \bibinfo{address}{New York, NY, USA}, \bibinfo{pages}{859–873}.
\newblock
\showISBNx{9781450371025}
\urldef\tempurl%
\url{https://doi.org/10.1145/3373376.3378508}
\showDOI{\tempurl}


\end{thebibliography}

\end{document}